\documentclass{article}

\usepackage{microtype}
\usepackage{graphicx}
\usepackage{booktabs} % for professional tables
\usepackage{arydshln}
\usepackage{hyperref}

\usepackage[accepted]{icml2024}

\usepackage{amsmath}
\usepackage{amssymb}
\usepackage{mathtools}
\usepackage{amsthm}
\usepackage[normalem]{ulem}

\usepackage[capitalize,noabbrev]{cleveref}

\usepackage{multirow}

\usepackage{import}
\usepackage[utf8]{inputenc} % allow utf-8 input
\usepackage[T1]{fontenc}    % use 8-bit T1 fonts
\usepackage{url}            % simple URL typesetting
\usepackage{placeins}            % simple URL typesetting
\usepackage{booktabs}       % professional-quality tables
\usepackage{amsfonts}       % blackboard math symbols
\usepackage{nicefrac}       % compact symbols for 1/2, etc.
\usepackage{microtype}      % microtypography
\usepackage{xcolor}         % colors

\usepackage{multicol}
\usepackage[hang,flushmargin]{footmisc} % Add footmisc package with appropriate options

\usepackage[font=footnotesize,labelfont=bf]{caption}
\usepackage{subcaption}
\usepackage{bm}
\usepackage{enumerate}
\usepackage{amsmath}
\usepackage{amssymb}
\usepackage{amsthm}
\usepackage{mathtools}
\usepackage{stmaryrd}
\usepackage{graphicx}
\usepackage{centernot}
\usepackage{gensymb}
\usepackage{marginnote}
\usepackage{wasysym}
\usepackage{physics}
\usepackage{empheq}
\usepackage{tikz}
\usepackage{qcircuit}
\usepackage{color, soul}
\usepackage{enumitem}
\usepackage{authblk}
\usepackage{framed}
\usepackage{comment}
\setlist[itemize,enumerate]{noitemsep, topsep=0pt, partopsep=0pt, parsep=0pt}

\usepackage[textsize=tiny]{todonotes}

\usetikzlibrary{positioning}
\usetikzlibrary{calc}
\usetikzlibrary{matrix}
\usetikzlibrary{decorations.pathreplacing,calligraphy}

\newtheorem{theorem}{Proposition}[section]

\newtheorem{lemma}[theorem]{Lemma}

\theoremstyle{remark}

\newcommand{\birkhoffp}[1]{\Omega_{#1}}				% Birkhoff polytope
\newcommand{\rowstochasticset}[1]{\Omega_{#1}^{(r)}}		% Set of row-stochastic matrices
\newcommand{\bvecnodim}[1]{\ensuremath{\mathbf{e}_{#1}}}	% Standard basis vector
\newcommand{\twodots}{\mathinner {\ldotp \ldotp}}
\newcommand{\intset}[2]{\ensuremath{\left[#1\twodots{} #2\right]}}
\newcommand{\uniopset}[1]{\ensuremath{\mathfrak{U}(#1)}}		% Set of unitary operators
\newcommand{\psimplex}[1]{\ensuremath{\Delta_{#1}}}		% Probability simplex
\newcommand{\vecmu}{\ensuremath{\boldsymbol{\mu}}}
\newcommand{\vecnu}{\ensuremath{\boldsymbol{\nu}}}
\newcommand{\tranpoly}[2]{\ensuremath{\mathcal{N}\left(#1, #2\right)}}

\DeclareMathOperator{\vecop}{\mathrm{vec}_r}		% Vec operator
\DeclareMathOperator{\traceop}{Tr}					% Trace
\DeclareMathOperator{\diagop}{diag}					% Diagonal
\DeclareMathOperator{\convop}{conv}					% Convex hull op
\newcommand{\idenm}[1]{\ensuremath{\mathbb{I}_{#1}}}		% Identity matrix I_n
\newcommand{\onesv}[1]{\ensuremath{\mathbf{1}_{#1}}}		% Vector of ones (n-dim)
\DeclarePairedDelimiter\ceil{\lceil}{\rceil}

\icmltitlerunning{Quantum Theory and Application of Contextual Optimal Transport}

\begin{document}

\twocolumn[
\icmltitle{
Quantum Theory and Application of Contextual Optimal Transport
}




\begin{icmlauthorlist}

\icmlauthor{Nicola Mariella}{ibm-dublin}
\icmlauthor{Albert Akhriev}{ibm-dublin}
\icmlauthor{Francesco Tacchino}{ibm-quantum}
\icmlauthor{Christa Zoufal}{ibm-quantum}
\icmlauthor{Juan Carlos Gonzalez-Espitia}{ibm-quantum,polimi}
\icmlauthor{Benedek Harsanyi}{ibm-zurich,epfl}
\icmlauthor{Eugene Koskin}{ibm-dublin,ucd}
\icmlauthor{Ivano Tavernelli}{ibm-quantum}
\icmlauthor{Stefan Woerner}{ibm-quantum}
\icmlauthor{Marianna Rapsomaniki}{ibm-zurich}
\icmlauthor{Sergiy Zhuk}{ibm-dublin}
\icmlauthor{Jannis Born}{ibm-zurich}

\icmlaffiliation{ibm-dublin}{IBM Quantum, IBM Research Europe - Dublin}
\icmlaffiliation{ibm-zurich}{IBM Research, IBM Research Europe - Zurich, Switzerland}
\icmlaffiliation{ibm-quantum}{IBM Quantum, IBM Research Europe - Zurich, Switzerland}
\icmlaffiliation{ucd}{University College Dublin, Ireland}
\icmlaffiliation{epfl}{École Polytechnique Fédérale de Lausanne, Lausanne, Switzerland}
\icmlaffiliation{polimi}{Politecnico di Milano, Milan, Italy}

\icmlcorrespondingauthor{}{nicola.mariella@ibm.com}
\icmlcorrespondingauthor{}{jab@zurich.ibm.com}

\end{icmlauthorlist}

\icmlkeywords{Machine Learning, ICML}

\vskip 0.3in
]



\printAffiliationsAndNotice{}  % leave blank if no need to mention equal contribution

\begin{abstract}
Optimal Transport (OT) has fueled machine learning (ML) across many domains.
When paired data measurements $(\boldsymbol{\mu}, \boldsymbol{\nu})$ are coupled to covariates, a challenging conditional distribution learning setting arises.
Existing approaches for learning a \textit{global} transport map parameterized through a potentially unseen context utilize Neural OT and largely rely on Brenier's theorem.
Here, we propose a first-of-its-kind quantum computing formulation for amortized optimization of contextualized transportation plans.
We exploit a direct link between doubly stochastic matrices and unitary operators thus unravelling a natural connection between OT and quantum computation.
We verify our method (QontOT) on synthetic and real data by predicting variations in cell type distributions conditioned on drug dosage.
Importantly we conduct a 24-qubit hardware experiment on a task challenging for classical computers and report a performance that cannot be matched with our classical neural OT approach.
In sum, this is a first step toward learning to predict contextualized transportation plans through quantum computing.
\end{abstract}

\section{Introduction}
    
Optimal transport (OT)~\cite{VillaniOTOldNew} provides a mathematical framework for finding transportation plans that minimize the cost of
moving resources from a source to a target distribution.
The cost is defined as a distance or a dissimilarity measure between the source and target points, and the OT plan aims to minimize this cost while satisfying certain constraints.
OT theory has found applications across several fields, including biology where it gained popularity in single-cell analysis, an area of research rich in problems of mapping cellular distributions across distinct states, timepoints, or spatial contexts~\cite{klein2023mapping}. Notable biological tasks are reconstructing cell evolution trajectories~\cite{schiebinger2019optimal}, predicting responses to therapeutic interventions~\cite{bunne2023learning,bunne2022supervised} %, inferring spatial and signaling relationships between cells~\cite{cang2020inferring} 
and aligning datasets across different omic modalities~\cite{cao2022unified}.
From the OT perspective, source and target distributions are measurements of biomolecules of single cells.
%with or without spatial/temporal resolution~\cite{efremova2020computational}.

\begin{figure*}[!htb]
    \centering
    \includegraphics[width=0.75\linewidth]{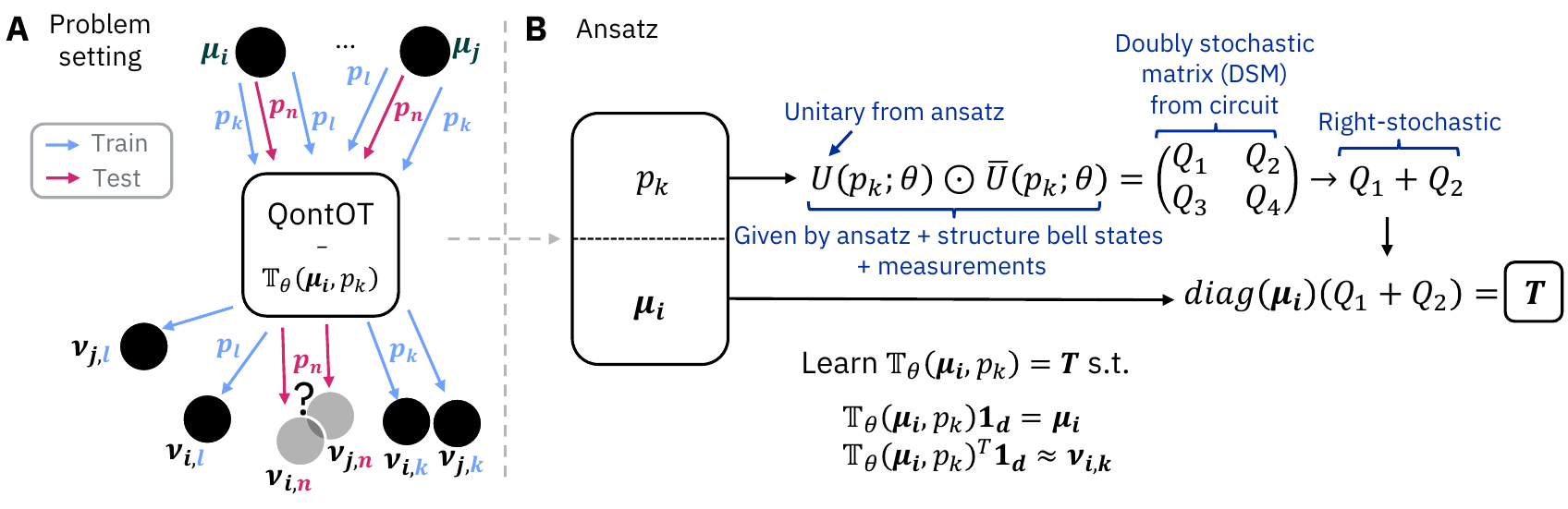}
    \caption{\textbf{A}) \textit{Contextual} OT is a conditional distribution learning problem. 
    \textbf{B}) Our proposed ansatz receives the context ($\boldsymbol{p}_k$) and the initial distribution $\boldsymbol{\mu}_i$ and produces a DSM that can be rescaled to a transport plan $T$ with marginal distributions $\boldsymbol{\mu}_i$ and $\boldsymbol{\hat{\nu}}_{i,k}$.
    }
    \label{fig:graphicalabstract}
    \vspace{-2mm}
\end{figure*}

In many OT applications, data measurements $\boldsymbol{\mu}_i$ (initial state) and $\boldsymbol{\nu}_i$ (final state) are coupled to a context $\mathbf{p}_i$ that induces $\boldsymbol{\mu}_i$ to develop into $\boldsymbol{\nu}_i$.
One thus might aspire to \textit{learn} a global transport map $T$ parameterized through $\mathbf{p}_i$ and thus facilitate the \textit{prediction} of target states
$\boldsymbol{\hat\nu}_j$ from source states $\boldsymbol{\mu}_j$, even for an unseen context $\mathbf{p}_j$ (cf.~\autoref{fig:graphicalabstract}A).
This work is largely based on Brenier's theorem~\yrcite{brenier1987decomposition} which postulates the existence of an unique OT map $T$ given by the gradient of a convex function, i.e., $T=\nabla f_\theta$.
\citet{makkuva2020optimal} showed that OT maps between two distributions can be learned through neural OT solvers by using a minimax optimization where $f_\theta$ is an input convex neural network (ICNN,~\citet{amos2017input}).
A notable example of such a neural OT approach is CondOT~\cite{bunne2022supervised} which estimates transport maps conditioned on a context variable and learned from quasi-probability distributions ($\boldsymbol{\mu}_i, \boldsymbol{\nu}_i$),
each linked to a context variable $\mathbf{p}_j$\footnote{
We use "contextual" rather than "conditional" to differentiate from OT on conditional 
    probabilities~\cite{tabak2021data}
}.
Limitations of such approaches include the dependence on squared Euclidean cost induced by Brenier's theorem~\cite{peyre2017computational} or the unstable training due to the min-max formulation in the dual objective as well as the architectural constraints induced by the partial ICNN.
% which is not always meaningful, especially for high-dimensional data.
The Monge Gap~\cite{uscidda2023monge}, an architecturally agnostic regularizer to estimate OT maps with any cost $C$, overcomes these challenges; however, unlike CondOT it cannot generalize to new context.
This is resolved in our concurrent work on the \textit{Conditional} Monge Gap~\cite{harsanyi2024learning}.

On a separate realm, quantum computing (QC) offers a new paradigm with the potential to become practically useful in ML~\cite{19HavSupervisedLearning, Liu_2021QuantumSpeedUpML, hhlPRL09, Huang21_powerofdatainQML} and fuel applications in, e.g., life sciences~\cite{basu2023towards} or high-energy physics~\cite{di2023quantum}.
A general hope of quantum ML lies in the identification of structures that can be produced more easily with quantum operators rather than classical principles.
Here, we propose a quantum contextual OT approach inspired by a natural link between OT and unitary operators, a fundamental concept in QC. 
This link is formed thanks to doubly stochastic matrices (DSMs). 
DSMs are real, nonnegative square matrices with row and column sums of 1.
They are a core structure behind the rise of OT in ML, for details we refer the reader to~\autoref{sec:linkOTandDSM}.
While ensuring the row \textit{or} column constraint is trivial (left- or right-stochastic matrices can be obtained with a simple softmax), producing DSMs parametrically is challenging with classical ML, typically iterative, non-parametric approximations like the Sinkhorn algorithm~(\citeyear{Sinkhorn1967ConcerningNM}) are used.
Thanks to the link of DSMs and unitary operators, we can turn the analytical problem of computing OT plans into a parameterizable approach to \textit{estimate} them.
In contrast to existing neural OT methods like CellOT~\cite{bunne2023learning}, our quantum formulation does not depend on Brenier's theorem (i.e., it is cost-agnostic) and unlike CellOT and the Monge Gap it estimates transportation plans explictly which is more interpretable, e.g. 
%, with the advantage of increased interpretability, e.g., 
the map topoloy can be studied.

Our key contributions can be summarized as follows:
\begin{enumerate}[leftmargin=*, labelindent=0pt]
    \item We are first to bridge QC with OT and ML. As shown in~\autoref{fig:graphicalabstract}\textbf{B}, we devise an ansatz that performs amortized optimization of contextual OT plans (or contextual DSMs) $Q \in \birkhoffp{d}$ given a context $\mathbf{p}_{\mathrm{new}}$ (\autoref{subsection-qc-birkhoff}). Given an unseen distribution $\vecmu_{\mathrm{new}}$, we extract a structure that can be rescaled to a transport map with a desired initial marginal distribution (\autoref{subsection-tmap-embedding}).
    \item We identify and leverage a previously unreported algebraic link between unitary operators and DSMs which connects OT and QC. 
    We prove that the constraints for a DSM can be obtained with quantum, yielding a "quantum inductive bias for DSMs".
    This is notable because it is currently unknown whether a similarly natural classical approach to parametrically produce DSMs exists.
    %equivalent to our "quantum inductive bias for DSMs" exists.
    \item We report a promising result on the relaxed assignment problem (i.e., contextual prediction of DSMs) obtained with 24-qubits on real hardware (IBM Quantum Platform) that outperforms a classical neural OT approach.
\end{enumerate}

The remaining paper begins with the contextual OT problem and proceeds with the quantum theory and the details of the ansatz (i.e., a parametric quantum circuit to approximate a quantum state)
for encoding DSMs and transport plans.
As a proof-of-concept, we first verify our method on synthetic and real drug perturbation data (with drug dosage as context).
% We use drug dosage as a context variable $\bm{p}_i$ and instantiate $\boldsymbol{\mu}_i$ and $\boldsymbol{\nu}_i$ as distributions over cell types for a cell population.
We then turn to a more constrained task, the contextual relaxed assignment problem which emphasizes better the strengths of our approach compared to classical neural OT.
% As our work describes a fundamentally novel approach to learn OT maps, our objective is to assess feasibility of learning contextualized transport plans through quantum, and \textit{not} to compete with established neural OT approaches.
% However, future extensions of our framework include approaches that directly predict the effect of a drug perturbation at the single-cell level.

\section{Preliminaries}
\subsection{Notation}
The sets of non-negative and positive reals are defined respectively as $\mathbb{R}_+$ and $\mathbb{R}_{++}$.
The $n$-dimensional vector of ones is denoted by $\onesv{n}$, the $n\times n$ identity matrix as $\idenm{n}$ and the $n\times n$ matrix of all ones as $J_n\coloneqq\onesv{n}\onesv{n}^\top$.
The set (with group structure) of unitary matrices of order $n$ is denoted $\uniopset{n}$.
Given a set $X$ we denote by $\convop(X)$ the convex hull of $X$, that is the minimal and unique convex set containing $X$.
We define the linear operator $\diagop(\cdot)$ as the mapping from a vector $\mathbf{v} \in \mathbb{C}^n$ to a diagonal matrix having as diagonal the same vector. 
The trace operator is denoted $\traceop(\cdot)$.
The notation $(\cdot)^{\dagger}$ is used to denote the complex conjugate transpose of the matrix or vector argument.
Let $A, B$ be matrices of the same size, then we denote by $A \odot B$ their \textit{Hadamard product}, which is defined as the
entry-wise multiplication $(A \odot B)_{i, j}=A_{i, j} B_{i, j}$. 
We use the \textit{Bra–ket notation} to denote the quantum states and their dual.
An integer symbol $i$ in the Ket $\ket{i}$, refers to the $i$-th basis vector in the computational basis.
Also, we make use of the following shortcut $\ket{ij}\coloneqq\ket{i}\otimes \ket{j}$.

\subsection{Linear Algebra of Doubly Stochastic Matrices}
\label{subsection-lin-algebra}
Let $\Delta_{n}$ denote the \textit{probability simplex} in $n - 1$ dimensions, 
%that is the set
\begin{align}
	\Delta_{n} =& \left\{\mathbf{v} \in \mathbb{R}_+^n\middle| \onesv{n}^\top\mathbf{v}=1\right\}.
\end{align}
Fundamental to this work is the \textit{Birkhoff polytope} $\birkhoffp{n}\coloneqq\mathcal{N}\left(\onesv{n}, \onesv{n}\right)$~\cite{brualdi_2006}, defined as the (convex) set of
$n \times n$ \textit{doubly stochastic matrices} (DSM).
A DSM $Q \in \birkhoffp{n}$ is a real, non-negative matrix with row/column sums of 1
\begin{align}
    Q\onesv{n} = \onesv{n},\quad&
    Q^\top\onesv{n} = \onesv{n},\quad Q_{i, j} \ge 0.
\end{align}
Permutation matrices are special cases of DSMs where the entries belong to $\{0, 1\}$.
Moreover, permutations are the only DSM that are unitary, so for any permutation $P$ we have that $PP^\top=P^\top P=\idenm{n}$.
A convex polytope is defined as the convex hull of a finite set of objects called extremes, 
in the case of the Birkhoff polytope, the extremes are the permutation matrices, this result is known as the \textit{Birkhoff theorem}.
Every DSM $Q \in \birkhoffp{n}$ can be decomposed as a convex combination of permutation matrices, that is
\begin{align}
	Q = \sum_{i=1}^{N} \lambda_i P_i,
\end{align}
for some probability vector $\boldsymbol{\lambda} \in \Delta_N$, $n\times n$ permutation matrices $\{P_i\}$, and the number of extreme points $N \le n^2$.
We note that the decomposition is \underline{not unique}.
Given a positive integer $n$, the number of permutation matrices is $n!$, which is the number of extreme points of $\birkhoffp{n}$.
However, the Birkhoff theorem bounds the number of permutations $N$ required to represent an arbitrary DSM by $n \cdot n$.

Another fundamental structure is the set $\mathfrak{S}_n$ (a subset of the Birkhoff polytope $\birkhoffp{n}$) of $n\times n$ \textit{unistochastic matrices}.
Given any $n\times n$ unitary matrix $U$, the matrix obtained by substituting each element of $U$ with its absolute value squared,
is unistochastic.
In other words, let $U \in \uniopset{n}$, then $U \odot \overline{U}$
is doubly stochastic, where $\overline{U}=\left(U^\dagger\right)^\top$. The latter result is an implication of unitarity.
The set of unistochastic matrices is a non-convex proper subset of the Birkhoff polytope, however the $n\times n$ permutations
matrices\footnote{
    A permutation matrix $P$ fulfills the unitarity constraints $P^\top P=PP^\top=\idenm{n}$, so $P \in \uniopset{n}$ (unitary).
    Also $P \odot \overline{P}=P$, then $P\in \mathfrak{S}_n$ (unistochastic).
} all belong to such set, hence its convex hull corresponds to the Birkhoff polytope,
that is $\convop\left(\mathfrak{S}_n\right)=\birkhoffp{n}$.
The constraints required for an arbitrary DSM to be unistochastic are still unknown\footnote{
    Unistochastic matrices of order 3 cover $\approx 75\%$ of the Birkhoff polytope~\cite{Dunkl_2009}.
}.

Let $\ket{b_n}$ denote the quantum state acting on $2n$ qubits\footnote{
    We consider Bell's states \cite{Nielsen} which are defined on a bipartite system and
    maximize the Von Neumann entanglement entropy.
}, consisting of $n$ maximally entagled states on $2$ qubits, so
% \footnote{
%    We motivate the adjective interleaved. Consider the following state from Bell's basis
%    $\ket{\phi_1}=\frac{1}{\sqrt{2}}\sum_{k=0}^1 \ket{k}\ket{k}$.
%    Then for $2n$ qubits we interleave $n$ Bell's states $\ket{\phi_1}$ to get
%    \begin{align*}
%        \frac{1}{\sqrt{2^n}}\sum_{k_1, k_2, \ldots=0}^1 (\ket{k_1}\ket{k_2}\cdots\ket{k_n})\otimes(\ket{k_1}\ket{k_2}\cdots),
%    \end{align*}
%    which corresponds to \eqref{eq:interleaved-bell-states-def}.
% }
%on
%$\left(\mathbb{C}^2\right)^{\otimes n} \otimes \left(\mathbb{C}^2\right)^{\otimes n}$, so
\begin{align}
    \label{eq:interleaved-bell-states-def}
    \nonumber
	\ket{b_n} =&
    \frac{1}{\sqrt{2^n}}\sum_{i_1,\ldots, i_n=0}^1 (\ket{i_1}\ket{i_2}\cdots\ket{i_n})\otimes(\ket{i_1}\ket{i_2}\cdots)\\
    =& \frac{1}{\sqrt{2^n}} \sum_{i=0}^{2^n - 1} \ket{i} \otimes \ket{i}.
\end{align}
We denote with $\vecop$ the \textit{row-major vectorization operator}.
Given a $n\times n$ matrix $M$ in $\mathbb{C}$, the latter operator is defined by the rule
\begin{align}
    \vecop(M) \coloneqq& \sum_{i=0}^{n - 1} M \ket{i} \otimes \ket{i}.
\end{align}
Moreover, we will be using the well-known identities linking vectorization to the Kronecker product
\footnote{In the context of quantum information theory the identity in Eq.~\eqref{vec-def} is known as the Choi-Jamio\l{}kowski correspondence.}
\begin{subequations}
\begin{align}
	\label{vec-def}
	\vecop(M)=&(M \otimes \idenm{n})\vecop\left(\idenm{n}\right)\\
	\label{vec-def-ii}
	\vecop\left(M^\top\right)=&(\idenm{n} \otimes M)\vecop\left(\idenm{n}\right).
\end{align}
\end{subequations}
We note that the state $\ket{b_n}$ in Eq.~\eqref{eq:interleaved-bell-states-def}, corresponds to the vectorization of the identity operator up to a scalar multiple, that is
$\frac{1}{\sqrt{2^n}} \vecop\left(\idenm{2}^{\otimes n}\right)=\ket{b_n}$.
The following lemma establishes a relation between unitary operators and their vectorization.
\begin{lemma}
	\label{lemma-u-to-vec-u}
	Let $\{U_i\}$ be a set of unitary operators $U_i \in \uniopset{n}$ such that
	$\traceop\left(U_i U_j^\dagger\right)=n \delta_{ij}$, that is the unitaries are
	orthogonal w.r.t. the Frobenius inner product.
	Then the set $\left\{\vecop(U_i)\right\} \subset \mathbb{C}^{n^2}$ consists of orthogonal vectors, that is
	\begin{align}
		\label{eq-lemma-u-to-vec-u}
		\vecop(U_j)^\dagger \vecop(U_i) =&
		\traceop\left(U_i U_j^\dagger\right)=n \delta_{ij}.
	\end{align}
\end{lemma}
Proof in Appendix \ref{appendix:proofs}.

\subsection{The Canonical OT Problem}
\label{section:intro-OT}
In the Kantorovich relaxation of the Monge problem \cite{peyre2017computational}, $C \in \mathbb{R}^{n\times m}$ is a non-negative matrix representing the cost of mass displacement
from entity $i$ to $j$ (so $C$ is called \textit{cost matrix} hereafter).
Let $\boldsymbol{\mu}, \boldsymbol{\nu}$ be strictly
positive real vectors (i.e., $\boldsymbol{\mu} \in \mathbb{R}_{++}^n$ and
$\boldsymbol{\nu} \in \mathbb{R}_{++}^m$), representing the quasi-probability discrete distributions\footnote{
    We say that a $d$-dimensional vector $\mathbf{v}$ is a quasi-probability discrete distribution when it is non-negative and non-zero.
    Then $\mathbf{v}/\left(\onesv{d}^\top \mathbf{v}\right)$ is a probability distribution.
} (also referred as states) for the source and destination entities.
The discrete (regularized) Kantorovich's OT problem is defined as
\begin{subequations}
\begin{align}
	\label{kantorovich-probl}
	\underset{Q \in \mathcal{N}(\boldsymbol{\mu}, \boldsymbol{\nu})}{\min}\,& \traceop\left(QC^\top\right) + \gamma h(Q),\\
	\text{s.t.}\quad&
	\mathcal{N}(\boldsymbol{\mu}, \boldsymbol{\nu}) =
	\left\{Q \in \mathbb{R}_+^{n \times m}\middle|
	Q\onesv{m}=\boldsymbol{\mu}, Q^\top\onesv{n}=\boldsymbol{\nu}\right\},
\end{align}
\label{eq:regKant}
\end{subequations}
where $h$ is a regularization function \cite{cuturi2013sinkhorn} with trade-off $\gamma \ge 0$.
The set $\mathcal{N}(\boldsymbol{\mu}, \boldsymbol{\nu})$ is the 
\textit{transportation polytope} \cite{brualdi_2006} whose elements
are \textit{transportation maps}.
Given a map $T \in \mathcal{N}(\boldsymbol{\mu}, \boldsymbol{\nu})$, $T_{i, j}$ represents the mass moved from source $i$ to destination $j$ (cf.~\autoref{fig:pictorial-tmaps}).
%In Figure \ref{fig:pictorial-tmaps} we propose a pictorial representation of the transportation maps.
As noted, the elements of the \textit{Birkhoff polytope}, a special case of the transportation polytope, are the DSMs which are the solutions to the
(relaxed) \textit{assignment problem}, where quasi-probability distributions
are uniform $\boldsymbol{\mu}=\boldsymbol{\nu}=\onesv{n}$.

\subsection{The Contextual OT Problem}
\label{section-ctx-ot-problem}
Neural OT is concerned with \textit{learning} the optimal transport between distributions from samples~\cite{makkuva2020optimal,korotin2023neural}, s.t. $\boldsymbol{\hat{\nu}}_i$ can be estimated from unseen $\boldsymbol{\mu}_i$.
Contextual OT generalizes this scenario: if $(\boldsymbol{\nu}_i, \boldsymbol{\mu}_i)$ are not observed in isolation but linked to a context $\mathbf{p}_i$, a conditional distribution learning task arises~\cite{bunne2022supervised,nguyen2024generative,harsanyi2024learning}.

Formally, let $\mathcal{K}_d\coloneqq\psimplex{d} \cap \mathbb{R}^d_{++}$ denote the subset of the probability simplex with vectors presenting non-zero components\footnote{To avoid degeneracy, Remark 2.1~\cite{peyre2017computational}.}.
We consider a dataset of contextualised measures each represented by a tuple
$(\mathbf{p}_i, (\boldsymbol{\mu}_i, \boldsymbol{\nu}_i)) \in \mathcal{X} \times \mathcal{K}_d^2$,
where the vector $\mathbf{p}_i \in \mathcal{X} \subseteq \mathbb{R}^s$ defines the context.
% The initial\footnote{The initial state is a control distribution in the context of single-cell perturbation responses.} and final states are represented by the vectors $\boldsymbol{\mu}_i$ and $\boldsymbol{\nu}_i$, respectively.
The initial and final states $\boldsymbol{\mu}_i$ and $\boldsymbol{\nu}_i$ are from the same set $\mathcal{K}_d$.
The cost matrix $C$ is not required to be constant across all samples, and can be interpreted as a materialization of the perturbation.
At inference time, we are given an unseen perturbation $\mathbf{p}_{\mathrm{new}}$ and initial state $\boldsymbol{\mu}_{\mathrm{new}}$,
and aim to predict a transportation map $T^\star \in \mathcal{N}(\boldsymbol{\mu}_{\mathrm{new}}, \boldsymbol{\nu}^\star)$,
s.t. marginalization yields the final states $\boldsymbol{\nu}^\star$.
At training time, we use classical OT solvers to obtain a map for each sample, providing a list of tuples $(\mathbf{p}_i, T_i)$ where $T_i \in \mathcal{N}(\boldsymbol{\mu}_i, \boldsymbol{\nu}_i)$ solves the $i$-th OT problem.

%As the first use case, we outline the \underline{conditional assignment problem}.
%This could be the ideal setting for the proposal in Section \ref{section-qot-formulation-i}.

%Let $C: \mathbb{R}^s \to \mathbb{R}_+^{d \times d}$ be a (stochastic)	% TODO Use p(C|p)
%mapping from a conditioning (perturbation) vector %(and possibly the initial state, when we consider an arbitrary polytope)
%to a cost matrix\footnote{In some literature \cite{comp-ot-peyre-cuturi} this is also known as the %"ground metric" (which is not necessarily a metric).}
%for the OT problem (denoted $C$ in \eqref{kantorovich-probl}).
%We observe realizations of $C$ as samples consisting of the tuples
%$\left\{(\mathbf{p}_i, C_i)\right\} \subset \mathbb{R}^s \times \mathbb{R}_+^{d \times d}$, that is %tuples of perturbation and corresponding ground metric.
%Using the training data we learn a function $f^*: \mathbb{R}^s \to \birkhoffp{d}$ from the sets of %conditioning vectors %(and possibly the initial state)
%to the optimal transportation plan in the polytope $\birkhoffp{d}$.
%The ideal function $f^*$ takes the following form
%\begin{align}
%	f^*(\mathbf{p}) =& \underset{Q \in %\birkhoffp{d}}{\mathrm{argmin}}\,\traceop\left(QC(\mathbf{p})^\top\right).
%\end{align}

%The generalization mechanism resides on a QNN that given a unobserved label $\mathbf{p}_{\mathrm{new}}$,
%and by using a proper ansatz,
%predicts a transportation plan without requiring the cost matrix $C$.

\section{Quantum Formulation}
\label{section-qot-formulation-i}

Our quantum formulation leverages the following fundamental concept.
Let $\overline{(\cdot)}$ denote the complex conjugate of the argument
(i.e. in the case of a matrix argument, the transpose of the adjoint),
and $\odot$ the Hadamard product between matrices.
If $U$ is a unitary matrix, then $U \odot \overline{U} \in \Omega_n$ is a DSM.
Hence we can represent (with some approximation) the solution of the assignment 
problem using unitary operators.
This principle produces DSM independently of the construction of the
unitary $U$, which offers great freedom in the choice of the ansatz for $U$ supporting both variational and possibly kernel-based learning.
Furthermore, such natural link between transportation maps and unitary operators may lead to quantum models enjoying better expressivity compared to classical counterparts~\cite{bowles2023contextuality,abbas2021power,liu2021rigorous,anschuetz2024arbitrary}.
For example, the multitask model by~\citet{bowles2023contextuality} corresponds to our DSM prediction which intrinsically possesses the required linear bias (i.e., all row and column sums equal 1).
%In fact, our problem can directly be related to recent results in quantum learning theory~\cite{bowles2023contextuality},
%where a separation between so-called quantum contextual and non-contextual models is proven under the assumption of specific constraints being present in the label space.
% Substituted the below with Francesco's suggestion above.
% It further reveals a natural link between transportation maps and unitary operators which could be a key component for an inductive bias
% that is well-suited for quantum computation \cite{bowles2023contextuality}.

\subsection{Quantum Circuit for the Birkhoff Polytope}
\label{subsection-qc-birkhoff}
\begin{figure}[!htb]
	\centering
    \begin{subfigure}[b]{0.48\columnwidth}
    	\begin{equation*}
    		\scalebox{0.6}{
    		\Qcircuit @C=0.5em @R=0.25em @!R {
        		\nghost{ {a}_{1} : \ket{0} } & \lstick{ {a}_{1} : \ket{0} } &
    			\qw  & \targ & \qw		& \qw                                              & \qw & \qw & \qw \\
    			\nghost{ {a}_{2} : \ket{0} } & \lstick{ {a}_{2} : \ket{0} } &
    			\qw  & \qw      & \targ	& \qw                                   		   & \qw & \qw & \qw \\
    			\nghost{ {a}_{3} : \ket{0} } & \lstick{ {a}_{3} : \ket{0} } &
    			\gate{\mathrm{H}}  & \ctrl{-2}    & \qw 		& \multigate{3}{U_p(\mathbf{p};\boldsymbol{\theta})} & \qw & \qw & \qw \\
    			\nghost{ {a}_{4} : \ket{0} } & \lstick{ {a}_{4} : \ket{0} } &
    			\gate{\mathrm{H}}  & \qw      & \ctrl{-2} 	& \ghost{U_p(\mathbf{p};\boldsymbol{\theta})}     & \qw & \qw & \qw \\
    			\nghost{ {j}_{1} : \ket{0} } & \lstick{ {j}_{1} : \ket{0} } &
    			\gate{\mathrm{H}} & \ctrl{2} & \qw		& \ghost{U_p(\mathbf{p};\boldsymbol{\theta})}        & \qw & \qw & \meter & \rstick{j_1}\\
    			\nghost{ {j}_{2} : \ket{0} } & \lstick{ {j}_{2} : \ket{0} } &
    			\gate{\mathrm{H}} & \qw      & \ctrl{2}	& \ghost{U_p(\mathbf{p};\boldsymbol{\theta})}		   & \qw & \qw & \meter & \rstick{j_2}\\
    			\nghost{ {i}_{1} : \ket{0} } & \lstick{ {i}_{1} : \ket{0} } &
    			\qw               & \targ    & \qw 		& \qw                                              & \qw & \qw & \meter & \rstick{i_1}\\
    			\nghost{ {i}_{2} : \ket{0} } & \lstick{ {i}_{2} : \ket{0} } &
    			\qw 			  & \qw      & \targ 	& \qw                               		       & \qw & \qw & \meter & \rstick{i_2}
    		}}
    	\end{equation*}
    	\caption{
     \autoref{subsection-qc-birkhoff}: DSM-encoding circuit}
        \label{fig-qc-dsm}
    \end{subfigure}
    % \hspace{1.5cm}
    \begin{subfigure}[b]{0.49\columnwidth}
    	\begin{equation*}
    		\scalebox{0.6}{
    		\Qcircuit @C=0.5em @R=0.25em @!R {
        		\nghost{ {a}_{1} : \ket{0} } & \lstick{ {a}_{1} : \ket{0} } &
    			\qw  & \targ & \qw		& \qw                                              & \qw & \qw & \qw \\
    			\nghost{ {a}_{2} : \ket{0} } & \lstick{ {a}_{2} : \ket{0} } &
    			\qw  & \qw      & \targ	& \qw                                   		   & \qw & \qw & \qw \\
    			\nghost{ {a}_{3} : \ket{0} } & \lstick{ {a}_{3} : \ket{0} } &
    			\gate{\mathrm{H}}  & \ctrl{-2}    & \qw 		& \multigate{3}{U_p(\mathbf{p};\boldsymbol{\theta})} & \qw & \qw & \qw \\
    			\nghost{ {a}_{4} : \ket{0} } & \lstick{ {a}_{4} : \ket{0} } &
    			\gate{\mathrm{H}}  & \qw      & \ctrl{-2} 	& \ghost{U_p(\mathbf{p};\boldsymbol{\theta})}     & \qw & \qw & \qw \\
    			\nghost{ i_1 : \ket{i_1} } & \lstick{ {i}_{1} : \ket{i_1} } &
    			\qw & \qw & \qw		& \ghost{U_p(\mathbf{p};\boldsymbol{\theta})}        & \qw & \qw & \meter & \rstick{j_1}\\
    			\nghost{ i_2 : \ket{0} } & \lstick{ {i}_{2} : \ket{0} } &
    			\qw               & \qw      & \qw   	& \ghost{U_p(\mathbf{p};\boldsymbol{\theta})}		   & \qw & \qw & \meter & \rstick{j_2}
    		}}
    	\end{equation*}
    	\caption{\autoref{subsection-tmap-embedding}: Embedded transport map}
        \label{fig-qc-rsm}
    \end{subfigure}
    \vspace{-1mm}
    \caption{
    Circuit structures for the transportation map prediction.
    The registers $\{i_k\} \cup \{j_k\}$ represent the bits for the index $(i, j)$ related to the entry $Q_{i, j}$ of the resulting DSM.
    The registry $\{a_k\}$ refers to the $2m$ auxiliary qubits as per~\autoref{subsection-qc-birkhoff}.
    Regarding~\autoref{fig-qc-dsm}, we remark that the registry $i$ has been added for construction reasons, however in practice it can be removed and substituted with
    a classical uniform sampling (using the computational basis states on $n$ qubits) over the registry $j$.
    Consequently, the number of required qubits for DSM-encoding can be reduced to $2m +n$.
    In~\autoref{fig-qc-rsm}, we have applied that trick to embed transportation maps, with $\ket{i_1} \in \{\ket{0}, \ket{1}\}$, as per~\autoref{subsection-tmap-embedding}.
    }
    \vspace{-6mm}
    \label{fig-qc-overall}
\end{figure}
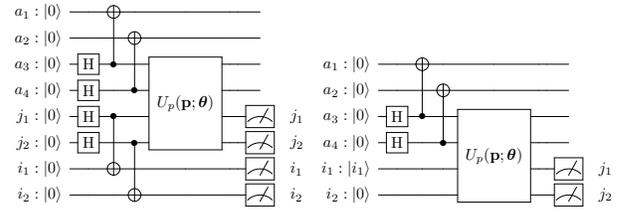

In this section, we assume that the part of the circuit acting on the first $n$ qubits has a dimension comparable to the input $d=2^n$,
where $d$ is the number of entities for the discrete distributions considered in~\autoref{section-ctx-ot-problem}.
Let $U_p$ be a parametric unitary operator acting on the
bipartite Hilbert space
$\left(\mathbb{C}^2\right)^{\otimes m} \otimes \left(\mathbb{C}^2\right)^{\otimes n}$,
and $m \in \mathbb{N}$ such that the classical simulation of a circuit on $m+n$ qubits is
intractable (with $m\ge n$) in general\footnote{
    We note that DSMs reside in a classical memory so they have a reduced space resources complexity.
    If we were not using $m$ auxiliary qubits for producing the DSM, then the circuit would
    require a logarithmic number of qubits w.r.t. the data size, so the unitary would have a reduced space complexity.
    Consequently, the resulting circuit would be tractable to classical simulation.
    % The additional $m$ qubits have other benefits that are revealed in the formulation.
}.
The operator $U_p$ depends on the input vector $\mathbf{p} \in \mathcal{X}$ (perturbation) as well as on the learning parameters $\boldsymbol{\theta}$.
To prove the construction, we consider the \textit{Operator-Schmidt decomposition} (\autoref{section:op-schmidt-decomp}) of $U_p$ (on $m+n$ qubits) determined by the quantum-mechanical
sub-systems $A_1, B_1$, consisting of respectively $m$ and $n$ qubits. 
%So
\begin{align}
	\label{op-schmidt-u_p}
	U_p(\mathbf{p}, \boldsymbol{\theta}) =& \sum_{i=1}^{d^2} \lambda_i V_i(\mathbf{p}, \boldsymbol{\theta}) \otimes W_i(\mathbf{p}, \boldsymbol{\theta})
\end{align}
with $\{V_i\}$ and $\{W_i\}$ being sets of unitary operators orthogonal w.r.t. the Frobenius inner product $\traceop\left(V_i V_j^\dagger\right)=2^m\delta_{ij}$.
The same follows similarly for the set $\{W_i\}$.
As a consequence of the SVD\footnote{The Operator-Schmidt decomposition is obtained through SVD.},
we have $\lambda_i \ge 0$, with unitarity of $U_p$ implying $\sum_i \lambda_i^2=1$.
Notably, the matrix $U_p$ depends on the input and the parameters vectors,
then the components of the Operator-Schmidt decomposition, namely $\lambda_i$, $V_i$ and $W_i$, are functions of $(\mathbf{p}, \boldsymbol{\theta})$.
Moreover, to assure the consistency of the formulation we impose the \textit{Schmidt rank} of $U_p$
(i.e. the number of strictly positive $\lambda_i$) to be greater than one\footnote{
    We impose the Schmidt rank for $U_p$ w.r.t. the split on $m+n$ qubits, to be $>1$.
    Otherwise the 
    partial trace, introduced on the $m$ qubits, makes the part of the unitary on $m$ qubits uninfluential.
}.
Using the unitary $U_p$ (omitting the dependency from $\mathbf{p}$ and $\boldsymbol{\theta}$ for clarity) and the states $\ket{b_n}$ and $\ket{b_m}$
(defined in Eq.~\eqref{eq:interleaved-bell-states-def}) we obtain the following state (on $2m + 2n$ qubits)
% \begin{subequations}
\begin{align}
	\label{varphi-def}
	\ket{\varphi} =& \left(\idenm{2}^{\otimes m} \otimes U_p \otimes \idenm{2}^{\otimes n}\right)\cdot\left(\ket{b_m} \otimes \ket{b_n}\right)\notag\\
	\underset{\eqref{op-schmidt-u_p}}{=}& \sum_k \lambda_k
	\left(\idenm{2}^{\otimes m} \otimes V_k\right)\ket{b_m}
	\otimes \left(W_k \otimes \idenm{2}^{\otimes n}\right)\ket{b_n}\notag\\
	\underset{\eqref{vec-def}}{=}&
	\sum_k \lambda_k 
	\frac{\vecop\left(V_k^\top\right)}{\sqrt{2^m}} 
	\otimes \frac{\vecop\left(W_k\right)}{\sqrt{2^n}},
\end{align}
% \end{subequations}
where $\vecop(\cdot)$ is the vectorization operator defined in~\autoref{subsection-lin-algebra}.
We note that the last equality is obtained using the identities in Eq.~\eqref{vec-def} and Eq.~\eqref{vec-def-ii}.
Now, we partition the Hilbert space on which $\ket{\varphi}$ lays into two subsystems. 
The first, $A_2$, consists of the first $2m$ qubits (auxiliary qubits).
The second, $B_2$, takes the last $2n$ qubits (data qubits)\footnote{We note that the systems $A_2$ and $B_2$ contain
respectively the systems $A_1$ and $B_1$, defined for the unitary in Eq.~\eqref{op-schmidt-u_p}.}.
We obtain the mixed state $\rho$ by applying the partial trace over the system $A_2$, to the pure state $\ket{\varphi}\bra{\varphi}$, that is
\begin{subequations}
\begin{align}
	\label{rho-def}
	\rho =& \traceop_{A_2}\left(\ket{\varphi}\bra{\varphi}\right)\\
	=& \frac{1}{2^{m+n}} \sum_{i, j} \lambda_i\lambda_j
	\traceop\left(V_i V_j^\dagger\right)
	\vecop\left(W_i\right) \vecop\left(W_j\right)^\dagger \notag \\
	\label{rho-def-decomp}
	\underset{\eqref{eq-lemma-u-to-vec-u}}{=}& \frac{1}{2^{n}} \sum_{i} \lambda_i^2
	\vecop\left(W_i\right) \vecop\left(W_i\right)^\dagger.
\end{align}
\end{subequations}
% \TODO{Note that the $\lambda_i$s depend on the $V_k$s and the parameters.
% Also stress that the number of summands depends on the size of the matrix (data size compatible).}
Recall that by the Operator-Schmidt decomposition and unitarity of $U_p$ we have that $\sum_i \lambda_i^2=1$.
Given that the action of the unitary $U_p(\mathbf{p}, \boldsymbol{\theta})$ is generally not classically efficiently simulable, the state $\rho$ has the potential to represent correlations that cannot be captured with classical models.
Moreover, here we can appreciate the role of the auxiliary $m$ qubits, that is enlarging the function space as a result of the convex combination of density matrices in Eq.~\eqref{rho-def-decomp}.
Indeed we note that if $m=0$, then the number of terms in Eq.~\eqref{rho-def-decomp} reduces to 1.
The recovery of the DSM is completed with the projective measurements explained in the lemma that follows (see resulting circuit in~\autoref{fig-qc-dsm}).
\begin{lemma}
	\label{lemma-transp-plan-recovery}
    Let
    \begin{subequations}
    \begin{align}
        \label{lemma:transp-plan-recovery-fun-p}
        p(i, j) \coloneqq& 2^n \traceop(\rho \ket{ij}\bra{ij})
    \end{align}
    for $i, j \in \intset{0}{d-1}$ and $\rho$ as defined in Eq.~\eqref{rho-def}.
    Let $\left\{\bvecnodim{i}\right\}$ be the set of canonical basis vectors (with index $i$ starting from 0) for the vector space $\mathbb{R}^{2^n}$.
	Then the matrix
    \begin{align}
        \label{lemma:transp-plan-recovery-final-q}
        Q=&\sum_{i, j=0}^{2^n - 1} p(i, j)\,\bvecnodim{i}\bvecnodim{j}^\top
    \end{align}
    \end{subequations}
    is doubly stochastic.
\end{lemma}
Proof in Appendix \ref{appendix:proofs}.
In the latter result, the rank 1 matrix $\bvecnodim{i}\bvecnodim{j}^\top$ corresponds to the $2^n\times 2^n$ matrix with $1$ in position $i, j$
and zeros elsewhere (i.e. the canonical basis for $2^n\times 2^n$ matrices in $\mathbb{R}$).
In other words, given the density matrix $\rho$ prepared as in Eq.~\eqref{rho-def}, the expectations w.r.t. the observables $\ket{ij}\bra{ij}$
provide the corresponding $(i, j)$ entry of the resulting matrix.
In practice, fixed an observable for entry $(i, j)$, we obtain a single bit of information for each execution of the circuit, so the resulting matrix
is guaranteed to fulfil the constraints of doubly stochasticity when the number of shots approaches infinity.
However, the convexity of the Birkhoff polytope offers great advantage in terms of restoring the DSM on a circuit,
more details are given in Appendix~\ref{section-dsm-estimation}. 
Our circuit is shown in~\autoref{fig-qc-dsm}.
\subsection{Embedding of Transportation Maps}
\label{subsection-tmap-embedding}
% The present section introduces an extension to Section \ref{subsection-qc-birkhoff}, that allows the encoding of generic transportation maps
% using quantum circuits. Moreover, as explained in Section \ref{section-ctx-ot-problem}, in the case of arbitrary transportation maps,
Since in our applications, the initial distribution $\vecmu$ is user-provided at inference time,
% \footnote{Specifically to the case in biology related to cell responses of therapeutical interventions, the initial distribution at inference time
% represents the characteristics of the tissue of an unobserved patient.}.
the problem is twofold; (1) embedding the transport map into a DSM to fit the representation presented in~\autoref{subsection-qc-birkhoff},
and (2) predicting maps which can be rescaled to an arbitrary initial distribution.
Given a data set $\left\{(\mathbf{p}_i, T_i)\right\}$ (i.e., tuples of contexts and transport maps), we assume $T_i \in \tranpoly{\vecmu_i}{\vecnu_i} \vecmu_i, \vecnu_i \in \mathbb{R}^d_{++}$, i.e., the margins of the transport maps are
strictly positive\footnote{
% The constraint on the strict positivity of the margins is justified in Section \ref{section-ctx-ot-problem}.
Justification on strict positivity of margins in~\autoref{section-ctx-ot-problem}.
}.
Let $\mathbf{v} \in \mathbb{R}^d$, and $D_\mathbf{v}=\diagop(\mathbf{v})$ the $d\times d$ diagonal matrix having the elements of
the vector $\mathbf{v}$ as diagonal elements.
Now, given $T_i$ as defined above, we define
% \begin{align}
$
    \widehat{T}_i \coloneqq D_{\vecmu_i}^{-1} T_i,
$
% \end{align}
and observe that $\widehat{T}_i \onesv{d}=D_{\vecmu_i}^{-1} T_i \onesv{d}=D_{\vecmu_i}^{-1} \vecmu_i=\onesv{d}$,
that is $\widehat{T}_i \in \tranpoly{\onesv{d}}{\vecnu_i^\prime}$ is a
right stochastic matrix\footnote{
% A non-negative $n \times n$ matrix $Q$ is right stochastic when the rows sum to one, that is $Q\onesv{n}=\onesv{n}$. \\
$Q \in \mathbb{R}_+^{n \times n}$ is right stochastic iff $Q \mathbf{1}_n = \mathbf{1}_n$. All DSMs are right stochastic but the converse is not true.
}, with
$\vecnu_i^\prime=\widehat{T}_i^\top\onesv{d}$. At inference, when given a perturbation $\mathbf{p}_i$, the model predicts a right stochastic matrix
$\widehat{T} \in \tranpoly{\onesv{d}}{\vecnu^\prime}$ for some $\vecnu^\prime \in \mathbb{R}^d_{++}$.
The latter, alongside the user-provided initial distribution $\vecmu$, determines the final predicted map $T=D_{\vecmu}\widehat{T}$, s.t. $T\onesv{d}=D_{\vecmu}\widehat{T}\onesv{d}=D_{\vecmu}\onesv{d}=\vecmu$.
In other words, we learn the transportation pattern in a margin-independent fashion and rescale to the required margin at inference time.
%\begin{remark}
   % \label{remark-null-ctx-mul-data-uploading}
Note that, when the context is $\mathbf{0}$ (null perturbation) then $\widehat{T}_i=\idenm{d}$.
Given some $\vecmu$ we have $D_{\vecmu}\widehat{T}_i=D_{\vecmu}$, hence $D_{\vecmu}\onesv{d}=D_{\vecmu}^\top\onesv{d}=\vecmu=\vecnu$, ergo the initial and final distributions agree (consistent with the notion of null perturbation), inducing a stationarity inductive bias. 

To confirm generality: any transport map $T \in \tranpoly{\vecmu}{\vecnu}$ with $\vecmu, \vecnu \in \mathbb{R}^d_{++}$,
can be decomposed as $T=DR$ where $R=\diagop(\vecmu)^{-1} T$ is right stochastic
and $D=\diagop(\vecmu)$ a positive diagonal\footnote{
    Indeed $R\onesv{d}=\diagop(\vecmu)^{-1} T\onesv{d}=\diagop(\vecmu)^{-1} \vecmu=\onesv{d}$ (right stochastic),
    so $DR=\diagop(\vecmu) \diagop(\vecmu)^{-1} T=T$.
}.
Conversely, the product $DR$ of a right stochastic $R$ and a positive diagonal $D$ (both of order $d$) is a transport map with
$\vecmu=D\onesv{d}$ and $\vecnu=(DR)^\top \onesv{d}$.

%
%
    % Assume that the context is a scalar $p \in \mathbb{R}$, then if the ansatz is characterized
    % by the fact that when its parameters $\boldsymbol{\theta}=\mathbf{0} \implies U=I$, then we can consider a multiplicative data uploading,
    % that is $\boldsymbol{\theta} \mapsto p\boldsymbol{\theta}$.
    % Interestingly, the \underline{Checkerboard Ansatz} introduced in Section \ref{subsection-checkboard-ansatz}, fulfils such requirement.
    % \TODO{Note this could be considered another (minor) inductive bias.}
%\end{remark}
%
\paragraph{Aggregation scheme.}
To complete the structure, we now expand on the link between the $d\times d$ right stochastic matrix $\widehat{T}$ and a $2d \times 2d$ DSM $Q$.
This step is necessary since the formulation in~\autoref{subsection-qc-birkhoff} produces only a DSM.
First, consider the DSM block decomposition 
\begin{align}
    \label{eq:dsm-partition-rsm}
    Q =&
    \begin{pmatrix}
        Q_1 & Q_2\\
        Q_3 & Q_4
    \end{pmatrix}
    \in \birkhoffp{2d},
\end{align}
with $Q_i \in \mathbb{R}_+^{d\times d}$.
Now, note that $\begin{pmatrix}Q_1 & Q_2\end{pmatrix}\onesv{2d}=\onesv{d}$ implies $\left(Q_1+Q_2\right)\onesv{d}=\onesv{d}$.
We embed the right stochastic matrix $\widehat{T}_i$ into the sum $Q_1+Q_2$ of the top quadrants of a DSM $Q\in \birkhoffp{2d}$.
Since this structure does not consider the submatrices $Q_3$ and $Q_4$, in Appendix~\ref{subsection-checkboard-ansatz}
we describe a custom designed ansatz that takes into account such invariant.
Moreover, as depicted in~\autoref{fig-qc-rsm}, we can obtain the sum $Q_1 + Q_2$ directly from the state preparation and
measurements. 
Specifically, by initialising the registry $j_2$ to $\ket{0}$ we obtain the top half of the matrix (w.r.t. rows)
and by tracing out the same registry we mimic the sum of the top two quadrants\footnote{see \textit{principle of implicit measurement} \cite{Nielsen}.}.
We call this "atop" aggregation.
To obtain the number of required qubits, let $m$ be the number of qubits (as per~\autoref{subsection-qc-birkhoff}) that makes the function space achievable by
the ansatz hard to be computed classically. Also, let $d=2^n$ and consider $d\times d$ transportation maps,
then using the reduction introduced in~\autoref{fig-qc-rsm},
the circuit requires $2(n+m+1)$ qubits\footnote{note that removing registry $i$ and using classical sampling of matrix rows could reduce to $(n+1) + 2m$ qubits} where $m\ge n+1$. 
In practice, we set $m:=n+1$ unless indicated otherwise.

% \TODO{(Optionally, expand the non-linear case for the initial distribution?)}

\subsection{Training Objective}
\label{sec:objective}
% We analyse the problem from the learning perspective.
Let $f: \mathcal{X} \to \rowstochasticset{d}$ be a function from the set of perturbations $\mathcal{X}$ to the set of $d\times d$ row-stochastic matrices,
and let $\mathcal{F}$ be the function space of such functions related to our model. Then, given the training set $\left\{(\mathbf{p}_i, T_i)\right\}$ we define
our learning problem via the loss
\begin{align}
    \mathcal{L}_T = & \underset{f \in \mathcal{F}}{\mathrm{min}}
    \sum_i \|D_{\vecmu_i}f(\mathbf{p}_i) - T_i\|_F^2,
    \label{eq:objective}
\end{align}
where $\vecmu_i=T_i\onesv{d}$ is the initial distribution for the $i$-th training sample and
$\|\cdot\|_F$ is the Frobenius matrix norm.
An optimal function in $\mathcal{F}$ that minimizes the loss $\mathcal{L}_T$ is denoted $f^{\star}$.
At inference time, given the initial (quasi-)distribution $\vecmu \in \mathbb{R}^d_{++}$ and the perturbation $\mathbf{p} \in \mathcal{X}$,
the predicted transportation map is obtained as $T=D_{\vecmu}f^\star(\mathbf{p})$.
Alternatively, the predicted target distribution $\boldsymbol{\nu}=\left(D_{\vecmu}f^{\star}(\mathbf{p})\right)^\top\onesv{d}$ can be directly optimized via:
\begin{equation}
    \mathcal{L}_M = 
    \underset{f \in \mathcal{F}}{\mathrm{min}}
    \sum_i \left\|\left(D_{\vecmu_i}f(\mathbf{p}_i)\right)^\top\onesv{d} - \bm{\nu}_i\right\|_2^2.
    \label{eq:objective_marginal}
\end{equation}
This can be interpreted as a weakly-supervised learning of transportation maps and is considered for comparison purposes.
The ansatz parameters are obtained via gradient-free optimization with COBYLA~\cite{powell1994direct}.

% BFGS~\cite{fletcher2000practical}; or, to reduce circuit evaluations, with faster, derivative-free alternatives like COBYLA~\cite{powell1994direct}.

%
\paragraph{Evaluation.}
Accuracy of transportation plan prediction is measured twofold.
First, the relative Frobenius norm
\begin{equation}
   F(\bar{T}_i, T_i) =  \frac{\|\bar{T}_i - T_i\|_F}{\|\bar{T}_i\|_F}
   \label{eq:frobenius}
\end{equation}
where $\bar{T}_i = D_{\vecmu_i}f^{\star}(\mathbf{p}_i)$
%, we report the relative norm because the absolute norm is unbounded \
Secondly, we report the sum of the absolute errors (SAE).
Accuracy of the predicted marginals $\bar{\vecnu}$ is measured through $L_2$ norm and $R^2$.

\subsection{Multidimensional OT}
This subsection shows how to estimate the bare minimum of necessary quantum resources, including even the case of discrete multidimensional OT~\cite{solomon2018optimal}, reflecting that many OT applications utilize multivariate rather than univariate measures (as assumed above).
% This subsection generalizes our framework to  \cite{solomon2018optimal} and shows how to estimate necessary quantum resources.
% Many OT applications involve multi-dimensional problems. The previous sections are developed under the assumption that the source and target sample spaces are at most countable, and the measures univariate. In this section we drop the second assumption.
Let the source data have $K$ covariates, i.e. $x_i =(x_i^{1}, \dots, x_i^{K})$. 
Assume that each covariate is defined on a discrete sample space $\mathbb{X}_k$ with cardinality $d^\mu_k$ and $d^\nu_k$ ($k \in [1,K]$) for source and target data, and define the probability space $\mathcal{P}_k=(\mathbb{X}_k, \mathcal{F}_k, \vecmu_k)$, where $\mathcal{F} = \sigma(\mathbb{X}_k)$ is the $\sigma$-algebra generated by $\mathbb{X}_k$, and $
\vecmu_k$ a measure on $(\mathbb{X}_k, \mathcal{F}_k)$. 
Then, the multi-dimensional source measure space is written as $\mathcal{P} = \left( \bigotimes_{k=1}^K \mathbb{X}_k, \bigotimes_{k=1}^K \mathcal{F}_k, \bigotimes_{k=1}^K \vecmu_k \right)$. 
Analogously, the target measure space is $\tilde{\mathcal{P}} = \left( \bigotimes_{l=1}^L \mathbb{Y}_l, \bigotimes_{l=1}^L \mathcal{G}_l, \bigotimes_{l=1}^L \vecnu_l\right)$.
With $K=L=1$, $d^\mu = P, d^\nu = R$, we recover the case discussed in \autoref{section:intro-OT}, i.e., $\vecmu$ and $\vecnu$ are vectors, and the transportation plan $T \in \mathbb{R}^{P \times R}$.
In general, $\boldsymbol{\mu} \in \bigotimes_{k=1}^K \mathbb{R}_{++}^{d_k^\mu}$ and $\boldsymbol{\nu} \in \bigotimes_{l=1}^L \mathbb{R}_{++}^{d_l^\nu}$,
%(the superscript is written to make emphasis on the cardinality of the corresponding state space). 
i.e., the size is governed by the state space cardinality.
Since the source and target distributions are represented by $K$ and $L$ rank tensors, the cost function will be a $(K+L)$-tensor.
Assuming identical source and target spaces with $K$ covariates, and $d$ states per covariate (i.e., $K=L$ and $d_k^\mu=d_k^\nu=d ~ \forall k \in [0,K]$) the OT plan spans $\mathcal{O}(d^{K})$ rows/columns. 
Notably, the discrete N-dimensional OT problem is $\# P$-hard~\cite{Bahar2023}.
%($\prod_{k}^N n_k \prod_{k}^M m_k$ in the general case).
Computing explicit OT plans thus quickly becomes demanding. 

Now consider the application by~\citet{bunne2022supervised} on predicting single-cell perturbation responses among $200$ cells with $50$ gene-based features. 
% \sout{$K=L=200$ cells with $d_k^\mu=d_l^\nu=50$ gene-based features per cell}.
In that case, $K=L=1$ and $d^\mu=d^\nu=50\cdot200$, hence our ansatz would require at least $56$ qubits (cf.~\autoref{subsection-tmap-embedding} bottom). 
% 28 qubits if d = 50
% \sout{Even considering only $K=L=20$ cells, $454$ qubits would be required.}
We prove in Appendix~\ref{section-dsm-estimation} that the minimal number of required shots $N_0$ in the right stochastic matrix scales in $\mathcal{O}(n \log n)$ with the number of rows/columns in the OT plan (see Eq.~\eqref{eq:error}).
% In this case $N_0>541$.
In this case $N_0>161$k.
With less shots the likelihood of empty rows in the matrix is high.
Furthermore to obtain a satisfactory sampling error for each entry we need $\mathcal{O}(d^2/\varepsilon^2)$ shots; for a precision $\varepsilon=0.01$ this is more than a trillion shots.
% Collecting more than $N_0$ is necessary for the algorithm to succeed as previously the probability of empty rows in the sampled matrix is high.
% Exceeding $N_0$ shots is necessary in practice for the algorithm's success, as otherwise the likelihood of empty rows in the matrix is high.
% \sout{However, in the simplified case of $20$ cells this is still intractable since $N_0 \approx 7\mathrm{e}{35}$.}
Even abandoning single cell resolution and setting $d^\mu=d^\nu=50$ still requires $28$ qubits, $N_0>541$ and $>25$M shots to obtain low error.

As a mitigation strategy, we cluster cells in our experiments into $d$ clusters and compute $\vecmu, \vecnu \in \mathbb{R}^d$, i.e., we set $K=L=1$ and $d$ to $8$ (or $16$) which requires $16$ (or $20$) qubits and $640$k ($2.56$M) shots.
% $N_0 \geq 72$ (or $155$) shots for $p=0.999$ (cf.~\autoref{eq:error}).
% Our method is therefore build in anticipation of growing hardware capabilities.
Note that prior art on neural OT for single-cell data~\cite{bunne2022supervised,bunne2023learning} 
%leverages ICNNs to recast the multidimensional OT problem to convex regression. 
optimizes over the push-forwarded measure so the OT plans can not be directly accessed, unlike in our quantum method.
\section{Experimental Setup}
\begin{figure}[!htb]
    \centering
    \includegraphics[width=1.01\linewidth]{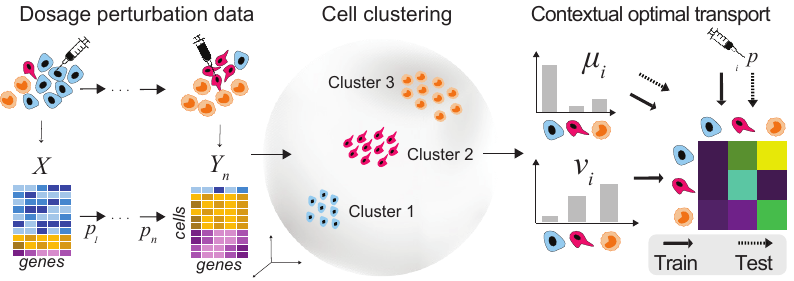}
    \caption{\textbf{Application overview}. 
    %We first employ a synthetic data generation scheme, where we perturb 
    A population of cells treated with varying drug dosages, resulting in ($X_i, Y_i, \bm{p}_i$) where $X_i$ ($Y_i$) represent scRNA-seq measurements before (after) a drug administered with dosage $\bm{p}_i \in [0, 1]$.
    We cluster the measurements to identify cell types and compute for each batch the distribution of cell types before and after perturbation, i.e., $\boldsymbol{\mu}$ and $\boldsymbol{\nu}$. 
    A classical OT solver computes the ground truth OT plan $T_i$ based on $\vecmu_i$, $\vecnu_i$ (not shown).
    Given our initial cluster distribution before perturbation $\boldsymbol{\mu}_i$ and the dosage $\bm{p}_i$ our ansatz predicts a transport plan $\bar{T}_i$.}
    \label{fig:overview}
    \vspace{-2mm}
\end{figure}
We applied our method on predicting changes in the composition of a cell population due to drug perturbations and tested it on synthetic and real data as shown in~\autoref{fig:overview}. 
Starting from a population of heterogeneous cells, each living in a high-dimensional state space we know that administering a drug has a direct effect on the composition of the cell population, by eliminating certain cell types or pushing some other cell types to proliferate. We denote as $\boldsymbol{\mu}$, $\boldsymbol{\nu}_i$ the cell type distribution of a cell population \textit{before} and \textit{after} the drug perturbation with a context variable $\bm{p}_i$, i.e., the drug dosage $\in [0, 1]$. 
We measure performance on unseen dosages for different data splitting strategies.

\subsection{Cell Type Assignment via Clustering}
%Because modeling distributional shifts in the original gene space through predicting full transportation maps is computationally too expensive
We represent each cell through a single label, obtained by clustering from the original $\mathbb{R}^l$ space into $d$ clusters (i.e., cell types).
We compute $\boldsymbol{\mu}, \boldsymbol{\nu} \in \mathbb{R}^d$ as distributions over clusters for the population of cells before and after perturbation. 
We cluster with $k$-Means and $k=d=8$ or $16$ to adhere with the circuit requirements, i.e., $d$ must be a power of $2$ (in general, if $\log_2{(k)} \not \in \mathbb{N}$, we pad and set $d := 2^{\ceil*{\log_2{(k)}}}$ and fix the transport plan to be diagonal for the padded entries).
We then solve Eq.~\eqref{eq:regKant} and compute the OT map between $\boldsymbol{\mu}$ and $\boldsymbol{\nu}$ with the Sinkhorn solver~\cite{cuturi2013sinkhorn}, which may be sped up through approximate solvers~\cite{haviv2024wasserstein}.
%from the~\texttt{POT} library~\cite{flamary2021pot}.
Repeating this procedure for all dosages yields a dataset $\{T_i, \bm{p}_i\}_{i=1}^N$ of transport plan-perturbation tuples, processed as described in~\autoref{sec:objective}.
The cost $C \in \mathbb{R}^{d \times d}$ is the Euclidean or cosine distance between centroids.
% An alternative approach to generate cell labels is to perform dimensionality reduction on the training cells and rasterize the obtained space into equidistant hypercubes. 
% Cells are then assigned to the hypercube they originate from and the cost matrix is computed using the $L_1$ distance of the centers of the hypercubes, scaled by dimension weights (e.g., eigenvalues of a PCA).
% We test this procedure with a Kernel PCA~\cite{scholkopf1997kernel} using a radial basis function kernel and two principal components.
%
%
\vspace{-1mm}
\section{Experimental Results}
This section verifies that QontOT (\textbf{Q}uantum C\textbf{ont}extual \textbf{O}ptimal \textbf{T}ransport) can learn to predict transportation maps \textit{contextualized} through a perturbation variable.
\subsection{Synthetic Data}
Leveraging the established sc-RNA-seq generator~\texttt{Splatter}~\cite{zappia2017splatter}, we devised a perturbation data generator that allows to control the number of generated cells, genes, cell types, perturbation functions and more.
We experiment with different perturbation functions to up-/down-regulate gene expression linearly or nonlinearly, different distance metrics, number of clusters and data splits.
Even though the perturbation functions are simple and only affect expression of a subset of cells and genes, the induced changes in cell type distribution are significant, locally continuous and nonlinear (cf.~\autoref{fig:dists}).
Details on the synthetic data generator and the used datasets are in Appendix~\ref{sec:syndatagenerator} and~\ref{sec:params}.

We compare QontOT to two baselines, \textit{Average} and \textit{Identity}. 
\textit{Average} always predicts the same transportation plan, obtained by solving the regularized OT problem (Eq.~\eqref{eq:regKant}) on \textit{all} training samples at once, disregarding the context. 
\textit{Identity} always predicts the identity OT plan, s.t., $\vecnu = \vecmu$.
The results in~\autoref{tab:syntab} show that QontOT outperforms both baselines in all cases by a wide margin.
\begin{table}[!htb]
\begin{subtable}{\columnwidth}
\centering
\resizebox{1\columnwidth}{!}{%
\begin{tabular}{ccc|cc|cc}
\multicolumn{3}{c|}{} & \multicolumn{2}{c|}{\textbf{OT Plan}} & \multicolumn{2}{|c}{\textbf{Marginals}} \\
\textbf{Dist.} & \textbf{Method} & $\mathcal{L}$  & \textbf{SAE ($\downarrow$)} & \textbf{Frob.} ($\downarrow$) & \textbf{$L_2$} ($\downarrow$) & \textbf{$R^2$} ($\uparrow$) \\ 
\hline \hline
$L_2$ & Identity & -- & 1.50 & 1.41 & 0.69 & 0.28 \\
$L_2$ & Average & -- & 1.07 & 0.79 & 0.52 & 0.27 \\
%$L_2$ & Softmax & -- & 0.21 & 0.18 & 0.09 & 0.95 \\
$L_2$ & QontOT & $\mathcal{L}_T$ & 0.97 & 0.70 & 0.45 & 0.56 \\
% $L_2$ & QontOT-12 & $\mathcal{L}_T$ & 0.98 & 0.72 & 0.45 & 0.53 \\ 
$L_2$ & QontOT & $\mathcal{L}_M$ & 0.97 & 0.79 & 0.41 & 0.55 \\
% $L_2$ & QontOT-12 & $\mathcal{L}_M$ & 1.02 & 0.81 & 0.42 & 0.54 \\ 
\hline
Cos. & Identity & -- & 1.67 & 1.50 & 0.69 & 0.29 \\
Cos. & Average & -- & 1.11 & 0.82 & 0.52 & 0.27 \\
%Cos. & Softmax & -- & 0.21 & 0.17 & 0.09 & 0.96 \\
Cos. & QontOT & $\mathcal{L}_T$ & 0.97 & 0.71 & 0.44 & 0.59 \\
% Cos. & QontOT-12 & $\mathcal{L}_T$ &0.99 & 0.71 & 0.45 & 0.59 \\
Cos. & QontOT & $\mathcal{L}_M$ & 1.10 & 0.86 & 0.40 & 0.59 \\
% Cos. & QontOT-12 & $\mathcal{L}_M$ & 1.06 & 0.83 & 0.42 & 0.60 \\
\hline
\end{tabular}
}
% \captionsetup{font=scriptsize}
\vspace{-1mm}
\caption{Recovering effect of linear perturbation for different distances.}
\end{subtable} 
\vspace{0mm} % adjust the space as needed
\begin{subtable}{\columnwidth}
\vspace{1mm} 
\centering
\resizebox{1\columnwidth}{!}{%
\begin{tabular}{ccc|cc|cc}
% \multicolumn{3}{c}{} & \multicolumn{2}{c}{\textbf{Transportation plan}} & \multicolumn{2}{c}{\textbf{Marginals}} \\
\textbf{Dist.} & \textbf{Method} & $\mathcal{L}$ & \textbf{SAE ($\downarrow$)} & \textbf{Frob.} ($\downarrow$) & \textbf{$L_2$} ($\downarrow$) & \textbf{$R^2$} ($\uparrow$) \\ 
\hline \hline
$L_2$ & Identity & -- & 1.22 & 1.19 & 0.50 & 0.45 \\
$L_2$ & Average & -- & 0.97 & 0.72 & 0.41 & 0.42 \\
% $L_2$ & Softmax & -- & 0.24 & 0.22 & 0.12 & 0.90 \\
$L_2$ & QontOT & $\mathcal{L}_T$ & 0.86 & 0.62 & 0.34 & 0.47 \\
$L_2$ & QontOT & $\mathcal{L}_M$ & 0.97 & 0.77 & 0.32 & 0.48 \\
\hline
\end{tabular}
}
% \captionsetup{font=scriptsize}
\vspace{-1mm}
\caption{More realistic and challenging scenario of nonlinear perturbations.}
\end{subtable}
\vspace{-5mm}
\caption{\textbf{Transportation plan prediction.}
Performance in predicting transportation plans for unseen dosages and linear (\textbf{a}) and nonlinear (\textbf{b}) perturbations; comparing QontOT to two baselines.
% Each block of three lines denotes one dataset, evaluated with two QontOT models and the baseline.
Different distance metrics were used to derive the cost matrix from the $k$-means centroids and both linear and non-linear perturbation effects were recovered.
% $\mathcal{L}$ denotes whether marginals ($\mathcal{L}_M$) or transport plans ($\mathcal{L}_T$) were explicitly optimized.
SAE denotes sum of absolute errors and Frobenius is the relative Frobenius norm. %(cf.~\autoref{eq:frobenius}).
% L refers to the number of layers in the ansatz, linear and nonliear perturbation to $g_{p_1}$ and $g_{p_2}$.
Means across three simulations are shown, full results in~\autoref{tab:performance}.
}
\label{tab:syntab}
\vspace{-2mm}
\end{table}
The two flavors of QontOT, $\mathcal{L}_T$ and $\mathcal{L}_M$ both have respective advantages; $\mathcal{L}_T$ models are explicitly trained on the transport plan.
They provide solutions with lower cost but instead the $\mathcal{L}_M$ models only optimize the marginal distribution and give typically better results in $L_2$ and $R^2$.
% Unlike related work~\cite{bunne2022supervised}, our method is not limited to squared Euclidean cost and works equally when computing the cost through e.g., euclidean or cosine distances of the centroids. 
Unlike related work~\cite{bunne2022supervised}, our method supports various costs like Euclidean or cosine distances of centroids, not just squared Euclidean.
% QontOT outperforms the baseline not only in the linear but also in the non-linear perturbation case, 
%While these experiments used a linear perturbation function $f_{p_1}$, the last block of~\autoref{tab:performance} evaluates QontOT on recovering variations in cluster ID distributions induced by non-linear perturbations.
 % arguably more realistic and challenging scenario, as $g_{p_2}$ induces strong up-regulation of genes with low initial expression but very low up-regulation if genes are already highly expressed. 
The exemplary real and predicted transportation plans in~\autoref{fig:panel1}A show that QontOT learns context-dependent shifts in cell type frequencies, by capturing the change in the distribution of cluster labels induced by the perturbation.  
% Assessing the relation between dosage and error (cf.~\autoref{fig:panel1}B) reveals that 
Predicting the effect of stronger perturbations (higher dosages) is more challenging (cf.~\autoref{fig:panel1}B). 
This is expected because in the control condition ($\bm{p}_i=0$), the cell type distribution remains identical, subject only to stochastic effects in data generation and batch assembly.

\paragraph{Circuit ablations.}
Next we sought to assess the robustness of QontOT to different configurations: the choice of ansatz, the optimizer, the number of layers and auxilliary qubits and the aggregation scheme to obtain a right stochastic matrix.
This time we used synthetic data with four cell groups (rather than one) which simulates more complex tissue and yields richer OT plans.
Overall, QontOT is robust to small alterations in circuit structure (cf.~\autoref{tab:quantablation}).
Even though many settings use a slightly larger computational budget, none of them improve consistently across metrics over the base configuration, validating the imposed inductive biases, e.g., our ansatz type and the "atop" aggregation.
For example, replacing our gradient-free optimizer (COBYLA) with another one (Nevergrad~\cite{bennet2021nevergrad}) yields identical results. 
%Instead, early experiments with BFGS had led to better performance but due to the challenges in applying it on quantum hardware, it was not further considered.
Extending the number of layers in the ansatz or adding more auxilliar qubits was generally found beneficial but also does not always improve performance (more results in~\autoref{tab:layerablation}).
%The baseline model, however, performs relatively well for mediocre dosages but struggles with extreme values.
Moreover, we find that in an OOD (out-of-distribution) setting where we kept the $10\%$ highest dosages out, the test error increases only mildly compared to the training error (cf.~\autoref{fig:extrapolation}).

Note that the embedding for transport maps with given initial distribution $\vecmu$ proposed in~\autoref{subsection-tmap-embedding}
%(excluding the assignment problem)
can be adapted to classical neural networks.
The resulting quantum-inspired algorithm, which we call NeuCOT, is described in~\autoref{section-classical-embedding} and performs explicit optimization of transport plans without relying on Brenier.
Unlike QontOT it cannot be applied to the contextual (relaxed) assignment problem of predicting DSMs.
While QontOT uses gradient-\textit{free} optimization,
the quantum-inspired approach can be trained conventionally with backpropagation and thus outperforms QontOT on the aforementioned datasets. 
Gradient-based, quasi-Newton optimization through BFGS substantially improved QontOT's performance in simulation but it is currently not amenable to quantum hardware.
Encouragingly even with gradient-free optimization there are cases where QontOT yields identical or slightly better performance e.g., if the Hamiltonian of the system is known (cf.~\autoref{tab:neucot}).

\subsection{SciPlex Data}
To facilitate comparison with prior art, we compared QontOT to CellOT~\cite{bunne2023learning} and CondOT~\cite{bunne2022supervised} on two drugs from the SciPlex dataset~\cite{srivatsan2020massively} each administered in four dosages.
For each of the dosages and the control condition, $20\%$ of cells were randomly held out for validation.
\begin{table}[!htb]
\begin{subtable}{\columnwidth}
\centering
\begin{tabular}{c|cc|cc}
\textbf{Method} & \textbf{SAE ($\downarrow$)} & \textbf{Frob.} ($\downarrow$) & \textbf{$L_2$} ($\downarrow$) & \textbf{$R^2$} ($\uparrow$) \\ 
\hline \hline
Identity & $1.10$ & $1.04$ & $0.18$ & $0.47$ \\
% Average & $1.03$ & $0.73$ & $0.15$ & $0.62$ \\
% Softmax & $0.27$ & $0.23$ & $0.10$ & $0.78$ \\
QontOT-$\mathcal{L}_T$  & $0.78$ & $0.61$ & $0.17$ & $0.49$ \\
QontOT-$\mathcal{L}_M$  & $0.92$ & $0.68$ & $0.16$ & $0.57$ \\
CellOT  & $0.46$ & $0.41$ & $0.17$ & $0.52$ \\
CellOT-homo  & $0.68$ & $0.60$ & $0.29$ & $0.37$ \\
CondOT  & $0.45$ & $0.40$ & $0.18$  & $0.56$  \\
\hline
\hline
\end{tabular}
\vspace{-1mm}
\caption{\textbf{Mocetinostat}
}
\label{tab:mocetinostat}
\end{subtable}
\vspace{-2mm}
\begin{subtable}{\columnwidth}
\centering
% \resizebox{1.0\columnwidth}{!}{%
\begin{tabular}{c|cc|cc}
\textbf{Method} & \textbf{SAE ($\downarrow$)} & \textbf{Frob.} ($\downarrow$) & \textbf{$L_2$} ($\downarrow$) & \textbf{$R^2$} ($\uparrow$) \\ 
\hline \hline
Identity & $1.10$ & $1.01$ & $0.28$ & $0.43$ \\
% Average & $1.23$ & $0.82$ & $0.28$ & $0.18$ \\
% Softmax & $0.26$ & $0.23$ & $0.10$ & $0.83$ \\
QontOT-$\mathcal{L}_T$  & $0.82$ & $0.60$ & $0.17$ & $0.49$ \\
QontOT-$\mathcal{L}_M$  & $0.97$ & $0.82$ & $0.24$ & $0.47$ \\
CellOT  & $0.44$ & $0.40$ & $0.18$ & $0.56$ \\
CellOT-homo  & $0.93$ & $0.80$ & $0.45$ & $0.27$ \\
CondOT  & $0.70$ & $0.55$ & $0.32$  & $0.49$  \\
\hline
\hline
\end{tabular}
% }
\label{tab:pracinostat}
\caption{\textbf{Pracinostat}}
\end{subtable}
\caption{
\textbf{SciPlex comparison.}
Means across three runs are shown.
}
\label{tab:sciplex}
\end{table}
\autoref{tab:sciplex} indicates that CellOT largely yields the best results, note however, that it is an unconditional model and five models were trained (one per condition) inducing an unfair comparison.
When aggregating data across conditions (CellOT-\textit{homo}), performance drops below the level of QontOT-$\mathcal{L}_T$.
%This is a testimony of the competitiveness of our quantum approach.
CondOT is an evolution of CellOT that leverages partial ICNNs (PICNNs) and can be parameterized by dosage. 
This yielded overall the best results on transportation plan metrics but on the marginal metrics $L_2$ and $R^2$, QontOT is on par or even superior.
Notably, in many applications, such marginal metrics are of higher importance.
They can be directly optimized in QontOT's $\mathcal{L}_M$ optimization mode which depicts a form of weakly supervised amortized optimization of transport plans that also overcomes the dependence on conventional OT solvers to compute training samples (cf.~\autoref{section-ctx-ot-problem}).
The duality of QontOT's optimization mode is visualized and explained further in~\autoref{fig:umap}. 

\subsection{Contextual Relaxed Assignment Problem}
\label{sec:hardware}
Since our ansatz naturally emits a DSM, it can be applied to the contextual relaxed assignment problem directly.
In this task we predict
DSMs rather than generic transport plans.
In this case, initial and final distributions are fixed, thus we do not need the structures described in \autoref{subsection-tmap-embedding}.
We hypothesized that this task lies closer to the heart of our ansatz (e.g. it does not ignore $Q_3$ and $Q_4$) and thus decided to challenge QontOT in a stress test on quantum hardware.

%We applied QontOT on synthetic data to contextually predict DSMs and compared it to our quantum-inspired NeuCOT approach.
%We used $24$ qubits ($6$ data, $18$ auxillar), a circuit with depth $50$ and $\sim70$ ECR gates and a dataset of fourty $8\times8$ DSMs (randomly split into train and test) obtained by sampling from our circuit with random parameters $\in \mathcal{U}(-0.8\pi, 0.8\pi)$.
%For efficiency, the circuit was split (cf. Appendix \ref{sec:adhoc-circ-depth-optim}) and the smallest physical circuit layout was picked after 50,000 transpilations with the standard Qiskit transpiler.
%Parameters were optimized for $300$ steps over $13$ days on a 127-qubit device (IBM Sherbrooke) available through the IBM Quantum Platform. 
%No error mitigation was performed.
%
We applied QontOT on synthetic data to contextually predict DSMs and compared it to the classical NeuCOT approach.
We used $24$ qubits ($6$ data, $18$ auxiliary), a circuit with depth $50$ and $\sim70$ ECR gates and a dataset of fourty $8\times8$ DSMs (randomly split into train and test with $20\%$ test data) obtained by sampling from our circuit with random parameters $\in \mathcal{U}(-0.8\pi, 0.8\pi)$.
For efficiency, the circuit was split (cf. Appendix \ref{sec:adhoc-circ-depth-optim}) and the smallest physical circuit layout was picked after $50,000$ transpilations with the standard Qiskit transpiler.
Parameters were optimized for $235$ steps over $13$ days on a $127$-qubit device (IBM Sherbrooke) available through the IBM Quantum Platform. 
Since the algorithm requires state sampling, no error mitigation was performed and we collected only $8192$ shots per iteration.
Further details about the setup can be found in~\autoref{app:hardware-simul}.

% We ran the hardware experiment on 127-qubit IBM Sherbrooke device. Appendix~\ref{app:hardware-simul} provides some details about the simulation.
%
%
\begin{figure}[!htb]
    \centering
    \includegraphics[width=1\linewidth]{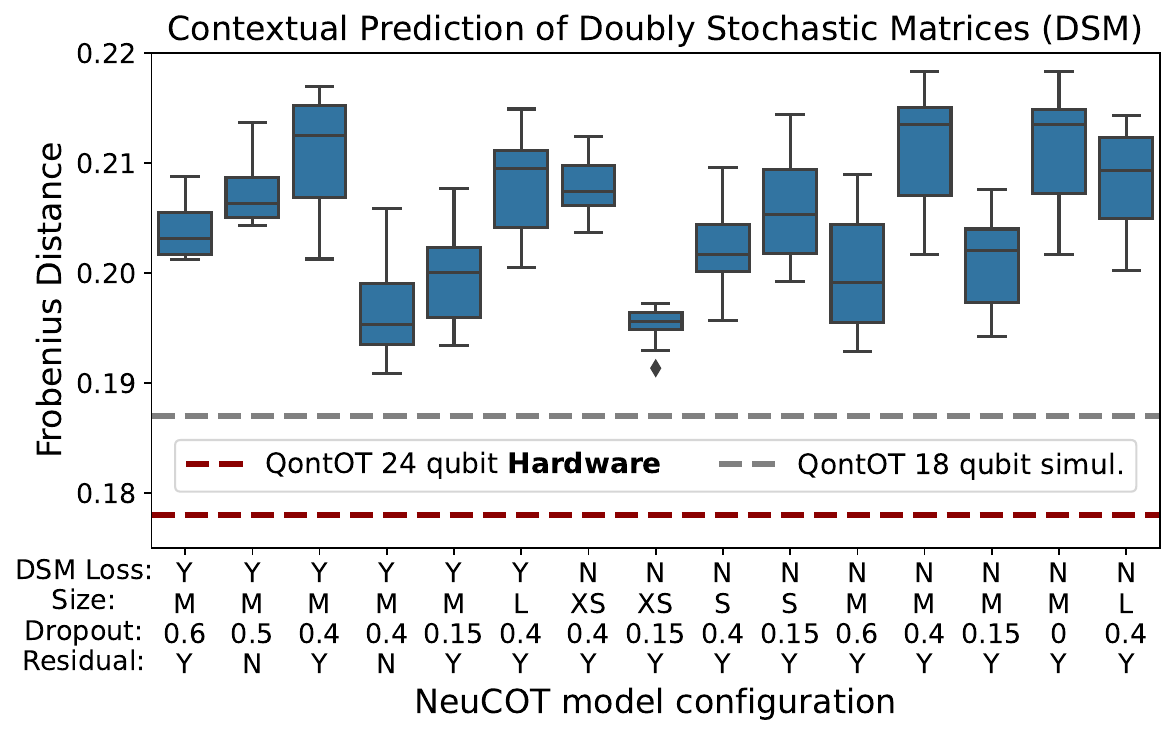}
    \vspace{-7mm}
    \caption{
    In a $24$-qubit hardware experiment, the performance of QontOT surpasses a $18$-qubit simulation and various classical neural OT models trained with different hyperparameter settings (cf.~\autoref{section-classical-embedding}).
    The NeuCOT models of size XS, S, M and L optimize respectively 3k, 8k, 81k and 5M parameters with ADAM compared to $124$ gradient-free optimized parameters in our ansatz.
    }
    \label{fig:hardware}
\end{figure}
The result shows that the objective function was converged well (cf.~\autoref{fig:fobj-profile-hw}). 
A few sudden spikes were observed during optimization; those are due to the device recalibration which was beyond our control.
The final result in~\autoref{fig:hardware} demonstrates that QontOT achieves a better Frobenius Distance than any of the $15$ tested NeuCOT configuration, each with different hyperparameters.
Interestingly, despite the device noise, the lack of error mitigation and the limited number of collected shots, the hardware run with a larger qubit budget ($24$) yielded better results than the $18$ qubit simulation.
Further performance metrics confirm this result.
%
%
\begin{comment}
\paragraph{Reconstructing gene expression}
Since cells are assigned unique cluster labels, QontOT predicts distributions over cluster assignments but this can not directly mapped back to single-cell gene expression.
As a remedy, one may utilize soft clustering such as fuzzy C-Means~\cite{bezdek2013pattern} followed by a propagation procedure to obtain predicted single-cell expression.
In this scenario, reconstructed gene expression may capture part of the response induced by the drug dosage (cf.~Appendix~\ref{sec:recon}).
% In~\autoref{fig:panel2}a, we verify that when predicting transportation maps for soft assignments, our results from~\autoref{fig:panel1} are confirmed.
% While the same general trends can be seen, an interesting observation is the high error of QontOT for low dosages in the Kernel PCA setting. 
% This error can be attributed to the stationarity inductive bias of the method (diagonal transportation plans for $p_i=0$).
% This is generally a desirable assumption of the method, but the kernelPCA seems to not always preserve local similarity of cells, inducing distribution shifts even for control states.
\end{comment}
%
\vspace{-6mm}
\section{Discussion}
%
% \begin{table}[!htb]
% \centering
% \resizebox{1.0\columnwidth}{!}{%
% \begin{tabular}{c|cc|cc}
% \multicolumn{1}{c}{} & \multicolumn{2}{c}{\textbf{Transportation plan}} & \multicolumn{2}{c}{\textbf{Marginals}} \\
% \textbf{Method} & \textbf{SAE ($\downarrow$)} & \textbf{Frob.} ($\downarrow$) & \textbf{$L_2$} ($\downarrow$) & \textbf{$R^2$} ($\uparrow$) \\ 
% \hline \hline
% Ident. & $0.45$ & $0.34$ & $0.00$ & $0.12$ \\
% Avg. & $1.32$ & $0.89$ & $0.00$ & $0.18$ \\
% Softmax & $0.05$ & $0.04$ & $0.00$ & $0.18$ \\
% QontOT-$\mathcal{L}_T$  & $0.27$ & $0.17$ & $0.00$ & $0.10$ \\
% % QontOT-$\mathcal{L}_M$  & $0.59$ & $0.43$ & $0.13$ & $0.58$ \\

% \hline
% \hline
% \end{tabular}
% }
% \caption{\textbf{Albert's DSM data - Extrapolation split}
% Used ansatz with 6 layers.
% }
% \label{tab:syn}
% \end{table}

Here, we introduced QontOT, a principled approach to represent transportation maps on quantum computers.
%Such representation is justified by the constraints defining the transportation maps, which can be related to recent work on
%quantum contextuality and inductive bias in quantum machine learning \cite{bowles2023contextuality}.
We proposed an ansatz for learning to predict OT maps conditioned on a context variable, without requiring access to the cost.
Our empirical results on synthetic and real data show that our method learns to predict contextualized transport maps which represent distributional shifts in cell type assignments.
% Exemplified on synthetic and real data of single-cell drug dosage perturbations, our method is able to predict transportation plans . 
While our method does not always match performance of the best classical models on this task, it constitutes, to the best of our knowledge, the first approach to bridge QC, OT and ML.
Notably, our approach does not impose constraints on the dimensionality of the context variable(s), thus more complex perturbations such as continuous drug representations, combinatorial genetic perturbations or other covariates could be employed.
However, given that the dosage-induced shifts in cluster assignments are also driven by the initial cell states (not only the dosage), future work could devise an ansatz fully parametric for $\vecmu_i$, potentially through (unbalanced) co-optimal transport~\cite{titouan2020co,tran2023unbalanced}.
Concurrently, we introduced a new classical neural OT baseline (NeuCOT) which may be further improved through a combination of neural and tensor networks~\cite{wang2023tensor} or a hybrid quantum-classical tensor network~\cite{schuhmacher2024hybrid}.

On the more constrained task of contextual prediction of DSMs we report a compelling finding from a noisy quantum computer, obtained without error mitigation and with gradient-free optimization whereas our classical competitor (NeuCOT) performs worse despite using backpropagation and orders of magnitudes more parameters.
Since we only used $24$ qubits, the physical circuit of that particular DSM prediction experiment may be reproduced by classical calculations.
In that sense, our method may be seen as a novel classical algorithm formulated in the language of quantum, however by applying it to larger sizes it could eventually lead to quantum advantage, assuming that the observed benefits are robust to scaling to more qubits.
Overall, our experiments suggest that predicting DSMs rather than generic transport maps is the more promising future endeavour.
A natural next step is to apply QontOT in scenarios where DSMs have to be estimated, one potential avenue could be Transformers where DSMs were found to emerge naturally~\cite{sander2022sinkformers}.
% and an inductive bias to enforce them yielded improved performance~\cite{sander2022sinkformers}.
\newpage
\section*{Impact Statement}
This paper presents work with the goal to advance the field of quantum machine learning (QML). 
There are many potential societal consequences of advances in quantum computing, first and foremost in cryptanalysis (see~\citet{scholten2024assessing} for a broad overview). 
Next, advances in QML will have more specific implications, e.g., they could widen disparities between organizations, countries or researchers that have access to, or can afford, leveraging quantum computing and those that cannot. 
More specific to our work, the promising result on contextual prediction of DSMs has to be seen in light of the limitations of current quantum hardware, e.g., we can only approximate specific states and structures (such as a DSM) since we have constraints in terms of circuit depth, the number of measurements and have to mitigate the device noise.
% However, beyond these general consequences, we can not weave any implications specific to our work.

\section*{Acknowledgements}
We thank the anonymous reviewers for their thorough feedback which helped to substantially improve the paper.

%that allows to map between different feature spaces.
% \TODO{Possible notes on topics/extensions non covered here: finite sampling of DSM, extension with non-linear action of $\vecmu$.}

% \section*{Software and Data}
% \section*{Acknowledgements}
% - Reviewers
% - Colleagues
% - Funding agencies
% - Sponsors


\bibliography{refs}
\bibliographystyle{icml2024}

%%%%%%%%%%%%%%%%%%%%%%%%%%%%%%%%%%%%%%%%%%%%%%%%%%%%%%%%%%%%%%%%%%%%%%%%%%%%%%%
%%%%%%%%%%%%%%%%%%%%%%%%%%%%%%%%%%%%%%%%%%%%%%%%%%%%%%%%%%%%%%%%%%%%%%%%%%%%%%%
% APPENDIX
%%%%%%%%%%%%%%%%%%%%%%%%%%%%%%%%%%%%%%%%%%%%%%%%%%%%%%%%%%%%%%%%%%%%%%%%%%%%%%%
%%%%%%%%%%%%%%%%%%%%%%%%%%%%%%%%%%%%%%%%%%%%%%%%%%%%%%%%%%%%%%%%%%%%%%%%%%%%%%%
\newpage
\appendix
\onecolumn
\section*{\Huge Appendix}
\appendix
 \renewcommand{\thefigure}{A\arabic{figure}}
\setcounter{figure}{0}
 \renewcommand{\thetable}{A\arabic{table}}
\setcounter{table}{0}

\section{Relation of DSMs and OT}
\label{sec:linkOTandDSM}
DSMs are a key structure within optimal transport because

\begin{itemize}
    \item They are at the root of the Kantorovich relaxation of the assignment problem. In the original assignment problem $n$ entities from the source distribution are assigned bijectively to $n$ entities from the target distribution through a mapping represented by a permutation matrix which is indeed invertible and orthogonal. In the Kantorovich relaxation the assignment takes a probabilistic form rather than a deterministic one (permutation matrices). This probabilistic form is represented through DSMs which are convex combinations of permutation matrices. More details can be found in the seminal textbook by~\citet{peyre2017computational} (see~\autoref{subsection-qc-birkhoff} and especially~\autoref{subsection-tmap-embedding}).
\item They were fundamental to formulate the entropically regularized version of OT~\cite{cuturi2013sinkhorn}. Cuturi's seminal work kicked off the integration of OT into modern ML. The discovery of the Sinkhorn divergence has been enabled through the Sinkhorn rescaling algorithm which converts nonnegative square matrices into DSMs by alternatively rescaling row and column sums~\cite{cuturi2013sinkhorn}. 
In essence, the prohibitively slow computation of OT (or earth-mover) distances — given by a linear program that requires super qubic runtime ($\mathcal{O}(n^3\log n)$, see~\citet{genevay2019sample}) — can be accelerated by an entropic regularisation term that converts the LP to Sinkhorn’s matrix rescaling algorithm.

\item DSMs can be rescaled to arbitrary transport maps, as demonstrated in~\autoref{subsection-tmap-embedding} of our paper. Thus they constitute a fundamental building block and can be leveraged not only for constrained applications where transport maps are exactly DSMs (e.g., in our experiment on quantum hardware on the ”contextual relaxed assignment problem”) but even for cases where the transport maps do not follow a specific structure (as shown in the remaining experiments of our paper).

\item DSMs can be linked to unitary operators thus connecting OT and quantum computing. This fundamental observation constitutes the foundation of our paper and is critical to address our specific task with quantum.

\end{itemize}

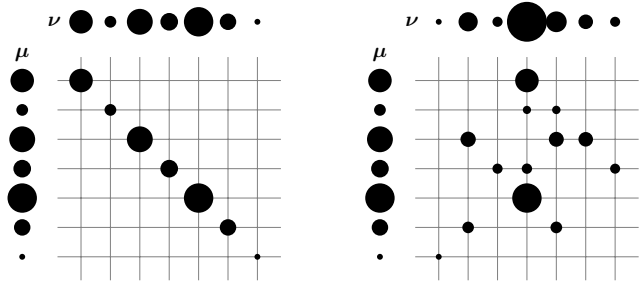
\begin{figure}[!htb]
\centering
\resizebox{0.5\columnwidth}{!}{%
\begin{subfigure}{0.49\columnwidth}
    \centering
    \resizebox{\linewidth}{!}{%
        \begin{tikzpicture}
    \newdimen\step
    \step=0.5cm
    \draw[step=\step, gray, thin] (-1.9cm,-1.9cm) grid (1.9cm,1.9cm);
    % Row margin
    \begin{scope}[xshift=-2.5cm, yshift=1.5cm, black]
        \foreach \x/\r in
            { 0/.2, 1/.1, 2/.22, 3/.15, 4/.25, 5/.14, 6/0.05 }
            \fill (0, -\x * \step) circle [radius=\r];
    \end{scope}
    % Column margin
    \begin{scope}[yshift=2.5cm, xshift=-1.5cm, black]
        \foreach \x/\r in
            { 0/.2, 1/.1, 2/.22, 3/.15, 4/.25, 5/.14, 6/0.05 }
            \fill (\x * \step, 0) circle [radius=\r];
    \end{scope}
    \node [anchor=south] at (-2.5cm, 1.7cm) {$\boldsymbol{\mu}$};
    % \node [anchor=east] at (-3.cm, 0) {$\boldsymbol{\mu}$};
    \node [anchor=south] at (-1.95cm, 2.3cm) {$\boldsymbol{\nu}$};

    \begin{scope}[yshift=1.5cm, xshift=-1.5cm, black]
        \foreach \x/\r in
            { 0/.2, 1/.1, 2/.22, 3/.15, 4/.25, 5/.14, 6/0.05 }
            \fill (\x * \step, -\x * \step) circle [radius=\r];
    \end{scope}
\end{tikzpicture}
    }
\end{subfigure}
\hspace{2cm}
\hfill
\begin{subfigure}{0.49\columnwidth}
    \centering
    \resizebox{\linewidth}{!}{%
        \begin{tikzpicture}
    \newdimen\step
    \step=0.5cm
    \draw[step=\step, gray, thin] (-1.9cm,-1.9cm) grid (1.9cm,1.9cm);
    % Row margin
    \begin{scope}[xshift=-2.5cm, yshift=1.5cm, black]
        \foreach \x/\r in
            { 0/.2, 1/.1, 2/.22, 3/.15, 4/.25, 5/.14, 6/0.05 }
            \fill (0, -\x * \step) circle [radius=\r];
    \end{scope}
    % Column margin
    \begin{scope}[yshift=2.5cm, xshift=-1.5cm, black]
        \foreach \x/\r in
            {0/0.05, 1/0.16330135, 2/0.08636299, 3/0.33930082, 4/0.17634963, 5/0.123911, 6/0.08541623}
            \fill (\x * \step, 0) circle [radius=\r];
    \end{scope}
    \node [anchor=south] at (-2.5cm, 1.7) {$\boldsymbol{\mu}$};
    \node [anchor=south] at (-1.95cm, 2.3cm) {$\boldsymbol{\nu}$};
    %\node at (-3.cm, 0) {$\mu$};
    %\node at (0, 3.cm) {$\nu$};

    \begin{scope}[yshift=1.5cm, xshift=-1.5cm, black]
        \foreach \i/\j/\r in
            {
            0/0/0.0,0/1/0.0,0/2/0.0,0/3/0.2,0/4/0.0,0/5/0.0,0/6/0.0,1/0/0.0,1/1/0.0,1/2/0.0,1/3/0.06985371860175908,1/4/0.07155737556329368,1/5/0.0,1/6/0.0,2/0/0.0,2/1/0.130177185610897,2/2/0.0,2/3/0.0,2/4/0.12688563490510568,2/5/0.12391100032343495,2/6/0.0,3/0/0.0,3/1/0.0,3/2/0.08636298980594585,3/3/0.08800853616657521,3/4/0.0,3/5/0.0,3/6/0.08541622535323554,4/0/0.0,4/1/0.0,4/2/0.0,4/3/0.25,4/4/0.0,4/5/0.0,4/6/0.0,5/0/0.0,5/1/0.09859630849369164,5/2/0.0,5/3/0.0,5/4/0.09939199138470259,5/5/0.0,5/6/0.0,6/0/0.05,6/1/0.0,6/2/0.0,6/3/0.0,6/4/0.0,6/5/0.0,6/6/0.0
            }
            \fill (\j * \step, -\i * \step) circle [radius=\r];
    \end{scope}
\end{tikzpicture}
    }
\end{subfigure}
}
    \caption{\textbf{Transportation maps}.
    The left and top sequences of blobs represent the initial ($\boldsymbol{\mu}$) and final ($\boldsymbol{\nu}$) distributions.
    The grid blobs denote the mass displaced from row $i$ to column $j$.
    The principle of mass preservation manifests as maintaining the total area of initial and final distribution blobs.
    % The principle of mass preservation materialises here as the preservation of the total area for the blobs related to the initial and final distributions.
    The left quadrant shows a diagonal transportation (without displacement) so $\boldsymbol{\mu}=\boldsymbol{\nu}$.}
    \label{fig:pictorial-tmaps}
    \vspace{-4mm}
\end{figure}

\section{The Operator-Schmidt decomposition}
\label{section:op-schmidt-decomp}
We start by defining the structure of the \textit{Schmidt decomposition} \cite{Bengtsson_Zyczkowski_2006}.
Let $\ket{\psi}$ denote a bipartite quantum state on the Hilbert space $\mathcal{H}=\mathcal{H}_1 \otimes \mathcal{H}_2$.
We assume the dimensions of $\mathcal{H}$, $\mathcal{H}_1$ and $\mathcal{H}_2$ are the positive integers $n$, $k_1$ and $k_2$, respectively, with $n=k_1 k_2$.
Then the Schmidt decomposition of $\ket{\psi}$ w.r.t. the split $\mathcal{H}_1 \otimes \mathcal{H}_2$ is defined as
\begin{align}
    \ket{\psi} \coloneqq \sum_{i=1}^{\min\{k_1, k_2\}} \lambda_i \ket{a_i} \otimes \ket{b_i},
\end{align}
where $\{\ket{a_i}\}$ and $\{\ket{b_i}\}$ are bases\footnote{
    The bases have cardinality respectively $k_1$ and $k_2$, however we note that the decomposition considers
    subspaces of dimension up to $\min\{k_1, k_2\}$.
} for respectively $\mathcal{H}_1$ and $\mathcal{H}_2$.
The coefficients $\lambda_i$, called \textit{Schmidt coefficients}, are real non-negative with $\sum_i \lambda_i^2 = 1$.
The Schmidt decomposition can be obtained through the singular value decomposition (SVD).
However, in our formulation we never compute such decomposition explicitly, instead, its formulation is used for proving the main results.
When a state decomposition is characterized by a single non-zero coefficient $\lambda_1=1$, we call it a \textit{product state}.
This is linked to a fundamental concept in quantum mechanics called \textit{entanglement}, and the product state represents the absence of it.

We now extend the decomposition to unitary operators. Let $U$ denote a unitary operator acting on $\mathbb{C}^n$.
Consider the following decomposition
\begin{align}
    \label{eq:vecu-schmidt-decomp}
    \vecop(U) = \sum_{i=1}^{\min\left\{k_1^2, k_2^2\right\}} \lambda_i \vecop(V_i) \otimes \vecop(W_i)
\end{align}
which can be interpreted as the Schmidt decomposition of the $n^2$-dimensional vector $\vecop(U)$.
Since the vectorization operator is an isomorphism, by inverting it we obtain
\begin{align}
    \label{eq:op-schmidt-decomp-def}
    U = \sum_{i=1}^{\min\{k_1^2, k_2^2\}} \lambda_i V_i \otimes W_i
\end{align}
which is the definition of Operator-Schmidt decomposition for $U$ w.r.t. the split $\mathcal{H}_1 \otimes \mathcal{H}_2$.
Finally, we note that Lemma \ref{lemma-u-to-vec-u} shows that the orthogonality of the vectors $\{\vecop(V_i)\}$ in \eqref{eq:vecu-schmidt-decomp},
corresponds to the orthogonality (w.r.t. the Frobenius inner product) of the operators $\{V_i\}$ in \eqref{eq:op-schmidt-decomp-def}.
Similarly, the same argument follows for $\{\vecop(W_i)\}$ and $\{W_i\}$.

\section{Quantum formulation}
\subsection{Recovery of finitely-sampled matrices}
\label{section-dsm-estimation}
In~\autoref{subsection-qc-birkhoff} we obtained a DSM from a quantum circuit by considering the measurements asymptotically.
The objective of the present section is that of obtaining a method for estimating the DSM from finite measurements.
We define a multiset as the tuple $\langle X, c_X\rangle$, where $X$ denotes the underlying set of elements and $c_X$ a function
mapping each element $x \in X$ to its cardinality.
Given the density matrix $\rho$ prepared as in Eq.~\eqref{rho-def}, we run a sampling process which produces a (nonempty) multiset $S=\langle \{(i, j)\}, c \rangle$
of pairs $\{(i, j)| i, j \in \intset{0}{d-1}\}$,
where $i$ and $j$ correspond respectively to the row and column indices of the DSM being sampled.
The pairs $(i, j)$ are counted using a (non-negative) $d\times d$ matrix $F$ whose entries $F_{i, j}$ correspond to the relative frequency of each pair,
that is
\begin{align}
    F_{i, j} =& \frac{c(\,(i, j)\,)}{\sum_{i^\prime, j^\prime} c(\,(i^\prime, j^\prime)\,)}.
\end{align}
By previous the definition we note that $\onesv{d}^\top F \onesv{d}=1$ (i.e. $F$ has total mass $1$),
and asymptotically the matrix $d F$ (i.e. rescaled to have total mass $d$) approaches the DSM in Eq.~\eqref{lemma:transp-plan-recovery-final-q}.
Assuming the matrix $F$ is non-zero (that is we acquired at least one sample), we define the projector onto the Birkhoff polytope as
\begin{align}
    \label{dsm-proj-probl}
    Q^\star =& \underset{Q \in \birkhoffp{d}}{\mathrm{argmin}} \left\|Q-dF\right\|^2_F.
\end{align}
Then, by the closedness and convexity of $\birkhoffp{d}$ the solution $Q^\star$ always exists and is unique.
We note that when $d F \in \birkhoffp{d}$, that is the input matrix is already DSM, then the minimizer becomes $Q^\star=d F$.
In other words, the projection acts as the identity operator when the input matrix belongs to the Birkhoff polytope.

\newcommand{\bb}[1]{\mathbf{#1}}
The remaining part of this section is going to focus on estimating the sampling error for the case of right stochastic matrices (see Section \ref{subsection-tmap-embedding})
obtained from the matrix of relative frequencies $F$.
To solve the problem, we employ the Kullback-Leibler (KL) divergence to quantify the informational difference between probability distributions.
% Let $F^\prime=\frac{d}{\onesv{d}^\top F \onesv{d}}F$ be the rescaled matrix of frequencies, such that the total mass \TODO{(motivate)} is $d$.
We minimize the KL divergence, so
\begin{subequations}
\begin{align}
    \underset{Q \in \mathbb{R}^{d\times d}_+}{\min}\,& \sum\limits_{i,j=1}^n Q_{i,j}\ln\left(\frac{Q_{i,j}}{d F_{i,j}}\right),\\
    \text{s.t.}\,& Q\onesv{d}=\onesv{d}.
\end{align}
\end{subequations}
Leveraging Lagrange multipliers, denoted as $a_i$, the objective function becomes
\begin{equation}
    \begin{aligned}
    \mathcal{L}(Q) &= \sum\limits_{i,j=1}^d\left( Q_{i,j}\ln\left(\frac{Q_{i,j}}{d F_{i,j}}\right)+a_i(Q_{i,j}-1) \right), \\
    \frac{\partial \mathcal{L}}{\partial Q_{i,j}} &= \ln Q_{i,j}-\ln(d F_{i,j})+1 + a_i=0 \implies
    Q=d \cdot \mathrm{diag}(\widetilde{\mathbf{a}})F,
    \end{aligned}
\end{equation}
with $\widetilde{\mathbf{a}} \in \mathbb{R}_{++}^d$.
By imposing the constraint $Q\onesv{d}=\onesv{d}$ and assuming that $F\onesv{d} \in \mathbb{R}_{++}^d$, it follows that
$\mathrm{diag}(\widetilde{\mathbf{a}})=\frac{1}{d}\mathrm{diag}(F \mathbf{1})^{-1}$, hence we obtain the minimizer
\begin{align}
    \label{eq:KL result}
    Q^\star &= d\cdot\mathrm{diag}(\widetilde{\mathbf{a}})F = \mathrm{diag}(F \mathbf{1})^{-1}F.
\end{align}

To obtain the latter we have assumed that $F\onesv{d} \in \mathbb{R}_{++}^d$, that is the row vectors of $F$ are non-zero.
To prevent this scenario, it is essential to determine the minimum number of shots required.
This count can be derived from the "Coupon Collector's Problem."~\cite{Blom1994}.
It indicates that to achieve a satisfactory probability (of obtaining $F$ with nonzero rows) $p$,
we need at least 
\begin{equation}
N_0=d\ln\left(\frac{d}{1-p} \right)
\label{eq:error}
\end{equation}
samples, and $1-p\ll1$.

We remark that the minimum number of samples $N_0$, relates to the requirement regarding the non-zero rows in the matrix $F$.
However, the latter does not cover the minimum number of shots to obtain a given precision $\varepsilon$ for each entry of the resulting matrix.
Indeed, for each entry of the matrix $Q^\star$ we need $\mathcal{O}(1/\varepsilon^2)$ measurements to obtain precision $\varepsilon$, consequently
we require $\mathcal{O}(d^2/\varepsilon^2)$ measurements for the entire matrix.

\subsection{The checkerboard ansatz}
\label{subsection-checkboard-ansatz}

We propose an ansatz construction which is convenient with respect to the structure of the embedding of transportation maps
expanded in Section \ref{subsection-tmap-embedding}. 
Specifically, since in the partitioning of the DSM in Eq.~\eqref{eq:dsm-partition-rsm}, only the top two quadrants contribute to the resulting
right stochastic matrix, we aim at devising an ansatz that does not carry additional information in the discarded quadrants.
The latter could also be interpreted as making the parametrisation for the ansatz more efficient.

Let $\sigma_i$ with $i=\intset{1}{3}$ be the Pauli operators commonly denoted with $\sigma_x$, $\sigma_y$ and $\sigma_z$, respectively.
Also we define $\sigma_0=\idenm{2}$. The subscript of the $\sigma$ to determine the Pauli will be indicated interchangeably as symbol or integer index.

For some positive integer $k$, we define the subset $G_k$ of unitary operators as 
\begin{align}
    \label{checkboard-set-def}
    G_k \coloneqq& \left\{U \in \uniopset{2k} \, \middle\rvert \,
    U = \idenm{2} \otimes A + \sigma_x \otimes B
    \right\},
\end{align}
where $A, B$ are $k\times k$ matrices, not necessarily unitary.
In other words, the operators in $U \in G_k$ have the following block matrix form
\begin{align}
    U =& \begin{pmatrix}
        A & B\\B & A
    \end{pmatrix},
    \label{eq:checkboard-op-pattern}
\end{align}
which is clearly inherited by the corresponding unistochastic
\begin{align}
    U \odot \overline{U}=&
    \begin{pmatrix}
    A \odot \overline{A} & B \odot \overline{B}\\
    B \odot \overline{B}& A \odot \overline{A}
    \end{pmatrix}
    =\begin{pmatrix}Q_1 & Q_2\\ Q_2 & Q_1\end{pmatrix}.
\end{align}
We now proceed with revealing the group-theoretical structure of the set $G_k$ and also its relation with the
tensor product, hence we obtain the construction of the ansatz implementing the unitary in Eq.~\eqref{eq:checkboard-op-pattern}.

The next lemma shows that the set $G_k$ is a subgroup of even degree of the unitary group.
\begin{lemma}
    \label{lemma-checkboard-group-struct}
    The set $G_k$ is non-empty and endowed with a group structure under operator composition,
    for all positive integers $k$.
\end{lemma}
\begin{proof}
    It is immediately verifiable that $\idenm{2k} \in G_k$, that is the set $G_k$ is non-empty and
    it contains the identity element w.r.t. matrix multiplication.
    Also the composition of operators carries the associativity as required.
    Finally we verify the closure.
    Let $U_1, U_2 \in G_{k}$ such that
    $U_i=\idenm{2} \otimes A_i + \sigma_x \otimes B_i$ for $i=1, 2$, then
    \begin{subequations}
    \begin{align}
        U_1 U_2 =&
        \left(\idenm{2} \otimes A_1 + \sigma_x \otimes B_1\right)
        \left(\idenm{2} \otimes A_2 + \sigma_x \otimes B_2\right)\\
        =& \idenm{2} \otimes \left(A_1 A_2 + B_1 B_2\right) +
        \sigma_x \otimes \left(A_1 B_2 + B_1 A_2\right),
    \end{align}
    \end{subequations}
    which corresponds to the pattern in \eqref{eq:checkboard-op-pattern}.
    Hence $U_1 U_2 \in G_k$.
\end{proof}

The result that follows shows that the structure is preserved under the tensor product.
\begin{lemma}
    \label{lemma-tensor-prod}
    Let $U_1 \in G_{k_1}$ and $U_2 \in G_{k_2}$, for some positive integers
    $k_1$ and $k_2$. Then $U_1 \otimes U_2 \in G_{2 k_1 k_2}$.
\end{lemma}
\begin{proof}
    Let $U_1 \in G_{k_1}$ and $U_2 \in G_{k_2}$ such that
    $U_i=\idenm{2} \otimes A_i + \sigma_x \otimes B_i$ for $i=1, 2$, then
    \begin{subequations}
    \begin{align}
        U_1 \otimes U_2 =&
        \left(\idenm{2} \otimes A_1 + \sigma_x \otimes B_1\right)
        \otimes \left(\idenm{2} \otimes A_2 + \sigma_x \otimes B_2\right)\\
        =& \idenm{2} \otimes A + \sigma_x \otimes B,
    \end{align}
    \end{subequations}
    with $A = A_1 \otimes \left(\idenm{2} \otimes A_2 + \sigma_x \otimes B_2\right)$
    and $B = B_1  \otimes \left(\idenm{2} \otimes A_2 + \sigma_x \otimes B_2\right)$.
    Hence it follows that $U_1 \otimes U_2$ fulfils the pattern in \eqref{eq:checkboard-op-pattern} and
    since $A, B$ are linear maps in $\mathbb{C}^{2k_1k_2}$, then $U_1 \otimes U_2 \in G_{2k_1k_2}$.
\end{proof}

\subsubsection{Ansatz's two-qubit generator}
We obtain a two-qubit circuit $U_g \in G_2$, that by Lemma \ref{lemma-checkboard-group-struct} and \ref{lemma-tensor-prod}
can be used as a generator for the more general $G_{2^k}$ with $k \ge 1$.
From the definition in Eq.~\eqref{checkboard-set-def} we obtain the symmetry
$U \in G_2 \implies (\sigma_x \otimes \idenm{2}) U (\sigma_x \otimes \idenm{2}) = U$.
Using the latter and the general unitary circuit with 2 CNOTs (highlighted) \cite{Barenco95}, we solve the following circuit equation 
\begin{equation}
    \label{ggen-circ-eq}
    \vcenter{
	\Qcircuit @C=0.5em @R=0.25em @!R {
		\nghost{ {q}_{0} } & \lstick{ {q}_{0} } &
		\qw  & \gate{C} & \targ     & \gate{R_z(\alpha)} & \targ     & \gate{A} & \qw &
		\qw & \qw &
		\gate{A^\dagger} & \targ     & \gate{R_z(-\alpha)} & \targ     & \gate{C^\dagger} & \qw &
		\qw & \qw
		\\
		\nghost{ {q}_{1} } & \lstick{ {q}_{1} } &
		\qw  & \gate{D} & \ctrl{-1} & \gate{R_y(\beta)}  & \ctrl{-1} & \gate{B} & \qw &
		\gate{X} & \qw &
		\gate{B^\dagger} & \ctrl{-1} & \gate{R_y(-\beta)}  & \ctrl{-1} & \gate{D^\dagger} & \qw &
		\gate{X} & \qw
		\gategroup{1}{4}{2}{8}{.7em}{--}
	}} = \idenm{2}^{\otimes 2},
\end{equation}
where $A, B, C, D$ are arbitrary single qubit (special) unitaries and $\alpha, \beta \in \mathbb{R}$.
We obtain a solution to the equation.
Since the operator $\sigma_z \otimes \idenm{2}$ commutes with the CNOT gate (with the Pauli $\sigma_z$ acting on CNOT's controlling qubit),
we impose on Eq.~\eqref{ggen-circ-eq} the conditions 
\begin{subequations}
\begin{align}
    B^\dagger \sigma_x B =& \sigma_z,\\
    R_y(-\beta) \sigma_z R_y(\beta) =& \sigma_z,\\
    D^\dagger \sigma_z D =& \sigma_x.
\end{align}
\end{subequations}
Then a solution is $B=D=H$ and $\beta=0$, where $H$ is the Hadamard operator on a single qubit.
Hence the generator circuit takes the following form
\begin{equation}
    \label{ggen-circ}
	\Qcircuit @C=0.5em @R=0.25em @!R {
		\nghost{ {q}_{0} } & \lstick{ {q}_{0} } &
		\qw  & \gate{C} & \targ     & \gate{R_z(\alpha)} & \targ     & \gate{A} & \qw\\
		\nghost{ {q}_{1} } & \lstick{ {q}_{1} } &
		\qw  & \gate{H} & \ctrl{-1} & \qw & \ctrl{-1} & \gate{H} & \qw
	}
\end{equation}
where $A, C \in SU(2)$ and $\alpha \in \mathbb{R}$.

\begin{figure}[h]
    \centering
    \begin{equation*}
        \Qcircuit @C=0.5em @R=0.50em @!R {
            \nghost{ {q}_{1} : \ket{0} } & \lstick{ {q}_{1} : } &
            \multigate{1}{\mathrm{K}(\boldsymbol{\theta}_1)} & \qw & \qw                                                & \qw\\
            \nghost{ {q}_{2} : \ket{0} } & \lstick{ {q}_{2} : } &
            \ghost{\mathrm{K}(\boldsymbol{\theta}_1)}        & \qw & \multigate{1}{\mathrm{K}(\boldsymbol{\theta}_3)}   & \qw\\
            \nghost{ {q}_{3} : \ket{0} } & \lstick{ {q}_{3} : } &
            \multigate{1}{\mathrm{K}(\boldsymbol{\theta}_2)} & \qw & \ghost{\mathrm{K}(\boldsymbol{\theta}_3)}          & \qw\\
            \nghost{ {q}_{4} : \ket{0} } & \lstick{ {q}_{4} : } &
            \ghost{\mathrm{K}(\boldsymbol{\theta}_2)}        & \qw & \multigate{1}{\mathrm{K}(\boldsymbol{\theta}_4)}   & \qw\\
            \nghost{ {q}_{5} : \ket{0} } & \lstick{ {q}_{5} : } &
            \qw                                              & \qw & \ghost{\mathrm{K}(\boldsymbol{\theta}_4)}          & \qw
        }
    \end{equation*}
    \caption{An example of depth and connectivity efficient (single) layer for the Checkerboard ansatz. Here the blocks $K$ correspond to the 2-qubits circuit
    in Eq.~\eqref{ggen-circ} and the vectors $\boldsymbol{\theta}_i$ are seven dimensional vectors parameterizing gates $A, C$ and $\mathrm{R}_z$ of Eq.~\eqref{ggen-circ}.}
    \label{fig:ansatz-layer}
\end{figure}
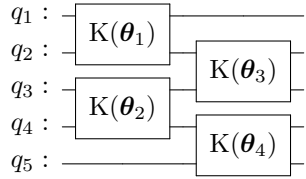

Finally, by using Lemma \ref{lemma-checkboard-group-struct} and \ref{lemma-tensor-prod}, and the generator block in Eq.~\eqref{ggen-circ},
we construct the ansatz as exemplified in~\autoref{fig:ansatz-layer}.

\subsection{Proofs}
\label{appendix:proofs}
\begin{proof}[Proof of Lemma \ref{lemma-u-to-vec-u}]
	Considering the constraint $\traceop\left(U_i U_j^\dagger\right)=n\delta_{ij}$ we obtain
	\begin{subequations}
	\begin{align}
		\vecop(U_j)^\dagger \vecop(U_i) =&
		\sum_k\left(\bra{k}U_j^\dagger \otimes \bra{k}\right)
		\sum_t\left(U_i\ket{t} \otimes \ket{t}\right)\\
		=& \sum_{k, t}\left(\bra{k}U_j^\dagger U_i\ket{t}\otimes \bra{k}\ket{t}\right)\\
		=& \sum_{k}\left(\bra{k}U_j^\dagger U_i\ket{k}\right)
		= \traceop\left(U_i U_j^\dagger\right)=n\delta_{ij}.
	\end{align}
	\end{subequations}
\end{proof}

\begin{proof}[Proof of Lemma \ref{lemma-transp-plan-recovery}]
    We expand the function $p: \intset{0}{d-1} \times \intset{0}{d-1} \to [0, 1]$ defined in \eqref{lemma:transp-plan-recovery-fun-p}, so
    \begin{subequations}
    \begin{align}
    	\label{proj-meas-ij-def}
    	p(i, j) \coloneqq& 2^n \traceop(\rho \ket{ij}\bra{ij})\\
    	=& \sum_{k} \lambda_k^2 \bra{ij}\vecop\left(W_k\right) \vecop\left(W_k\right)^\dagger \ket{ij}\\
    	=& \sum_{k} \lambda_k^2 \bra{i}\left(W_k \odot \overline{W_k}\right) \ket{j}.
    \end{align}
    \end{subequations}
	The positivity of the entries of the DSM is clear from the definition of $p(i, j)$.
	We prove the rows sum constraint for $Q=\sum_{i, j} p(i, j) \ket{i}\bra{j}$, that is
	\begin{subequations}
	\begin{align}
		Q\onesv{d} =& \sum_{i, j} p(i, j) \ket{i}\\
		=& \sum_{i} \ket{i} \cdot\left(\sum_{k} \lambda_k^2 \bra{i}\sum_j \left(W_k \odot \overline{W_k}\right) \ket{j} \right),
	\end{align}
	\end{subequations}
	where the rightmost sum equals the vector \onesv{m} since $W_k \odot \overline{W_k}$ is unistochastic,
	also $\sum_k \lambda_k^2=1$ (following from \eqref{op-schmidt-u_p}), hence $Q\onesv{d}=\onesv{d}$.
	Similarly the same holds for the columns sum constraint, hence the claim follows.
\end{proof}

\section{Experimental details}

\subsection{Synthetic dosage perturbation data generator}
\label{sec:syndatagenerator}
We leverage the established single-cell RNA sequencing generator \texttt{Splatter}~\cite{zappia2017splatter} to form a three-stage generator for drug dosage perturbation datasets:
\begin{enumerate}[leftmargin=*]
\item First, \texttt{Splatter} samples raw expression counts ($X \in \mathbb{R}^{n_1 \times l}$, with $n_1$ cells and $l$ genes) from zero-inflated negative binomial distributions (one per gene).
Sufficient statistics of all underlying distributions (Poisson, Gamma, Chi-Square) can be controlled.
\item We aim to produce a tuple of ($X_i, Y_i, \bm{p}_i$) where $X_i$ holds unperturbed base states of $n_1$ cells and $Y_i \in \mathbb{R}^{n_2 \times l}$ holds perturbed states of $n_2$ cells, resulting from a drug perturbation administered with dosage $\bm{p}_i \in [0, 1]$.
To derive the perturbed states $Y_i$, new base states $\bar{Y}_i$ are sampled with the same configuration used to generate $X_i$, mimicking that cells are being destroyed during measurement. 
Subsequently $Y_i = g(\bar{Y}_i, \bm{p}_i)$ where $g$ is the total effect on the cells, governed by a combination of noise terms and the immediate effect $g_p(\cdot)$ of the perturbation.
We assume that only $15\%$ of the genes alter their expression upon perturbation. 
In this case, we apply $g_p$ to the raw cell states, scaled by a response amplitude $\sim \mathcal{U}(0.3, 1)$.
Moreover, $10\%$ of the cells are generally unresponsive to the perturbation ($g_p=0$).
% We set $g_p(\cdot)=0$ for some unresponsive cells and most unresponsive genes, but for the fraction of responding genes ($15\%$ in our experiments) a perturbation effect $g_d$ is applied to the raw cell measurements, scaled by a response amplitude $\sim \mathcal{U}(0.3, 1)$, and then added to the base state.
We investigate linear and non-linear perturbations, i.e., $g_{p_1}(x) = ax+b$ and a reciprocal root function $g_{p_2}(x) = a x^{-b}$ with $a,b>0$, respectively.
The hyperparameters of the experiments can be found in Appendix~\ref{sec:params}.
\item We repeat stage 2 for each dosage by varying smoothly the immediate effect $g_p(\cdot)$ based on $\bm{p}_i$, resulting in a dataset $\{X_i, Y_i, \bm{p}_i\}_{i=1}^N $ of $N$ tuples.
Responsive genes are fixed across samples. The base states $X_i$ are \textit{identical} across all samples of the dataset, mimicking the common experimental setting where only one control population was measured~\cite{srivatsan2020massively}.
\end{enumerate}
\begin{figure}[!htb]
    \centering
    \includegraphics[width=1\linewidth]{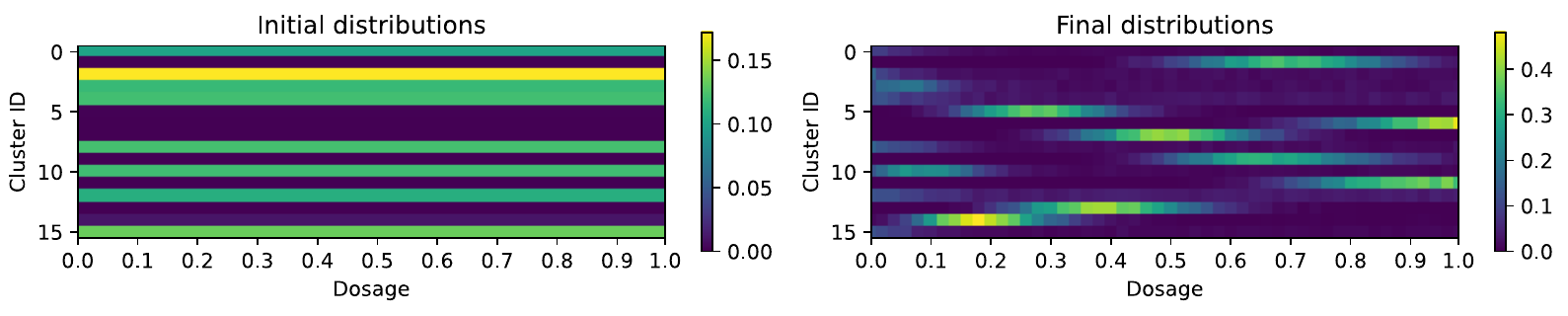}
    \caption{
    Exemplary cell type distributions for source and target cell populations.
    The distribution of cell types in the unperturbed tissue is either entirely static (see left) or varies mildly based on the user-defined noise level (not shown).
    Instead, the perturbed cells produce highly dissimilar distributions that, however, exhibit some local continuity for similar dosages.
    The plot was generated with data from linear perturbations and euclidean cost.
    }
    \label{fig:dists}
\end{figure}

\subsection{Datasets and hyperparameters}
\label{sec:params}
\subsubsection{Synthetic drug dosage perturbation data}
\begin{figure}[!htb]
    \centering
    \includegraphics[width=1\linewidth]{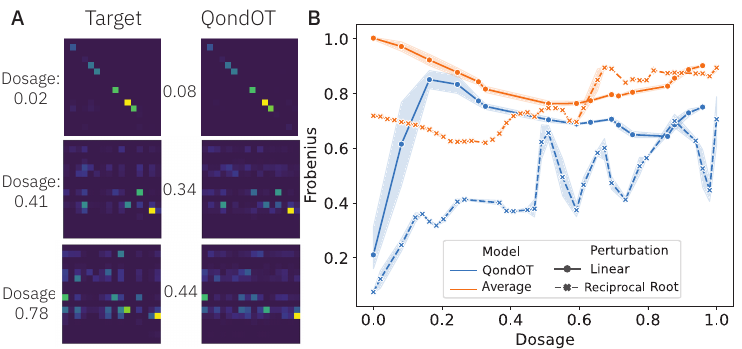}
    \caption{
    \textbf{Capturing variation in cell type distributions.}
    \textbf{a)} Three predicted transportation plans from the nonlinear perturbation dataset are shown next to their unseen ground truth.  
    \textbf{b)} Frobenius distance of real and predicted transportation across unseen dosages are shown for QontOT and the baseline.
    }
    \label{fig:panel1}
\end{figure}

\paragraph{Initial experiments}
For the results shown in~\autoref{tab:syntab} and~\autoref{fig:panel1}, we simulate 300 genes and 1000 cells across 50 unique dosages, equidistantly spaced in $[0, 1]$.
15\% of the genes respond to the perturbation $g_{p}(\cdot)$ but $10\%$ of the cells are set as unresponsive. The sinkhorn regularization $\gamma=0.001$.
For the linear case, $f_{p_1}(x) = 3x + 1$ and for the non-linear case  $f_{p_2}(x) = 100 x^{-0.2}$.
For each dosage, four batches of $500$ cells each were created, summing to 200 samples which were split randomly with $20\%$ held out dosages for testing.
In almost all experiments we set the number of clusters to $k=d=8$; only in~\autoref{fig:panel1} it was set to $16$.

\paragraph{Four cell groups.}
For the ablation studies on circuit structure (cf.~\autoref{tab:quantablation}), we simulate populations of $100$ genes and $2000$ cells, each belonging now to one out of four groups to simulate more complex tissue.
We perturb the populations with the function $f_{p_2}$ (as above) and $100$ dosages, equidistantly spaced in $[0,1]$.
The initial cell states $X_i$ are resampled for every dosage and only $2\%$ of cells are set as unresponsive.
The $100$ dosages are split randomly with $20\%$ test data.

\subsubsection{SciPlex data}
\label{sec:sciplex}
Selected models were trained on two of the nine compounds from the SciPlex dataset~\cite{srivatsan2020massively}.
High Throughput Screens on three cell lines were conducted with four varying concentrations (10, 100, 1000, and 10000 nM) for each drug.
We inherit preprocessing from~\citet{bunne2023learning} and~\citet{lotfollahi2019scgen} which includes library size normalization, filtering at the cell and gene levels, and log1p transformation. 
For mocetinostat and pracinostat we obtained respectively 22,154 and 21,926 cells from which 17,565 and 15,137 were control cells.
Data was split per condition (control + four dosages) in a roughly 80/20 ratio.
Preprocessing identified 1,000 highly-variable genes, which were compressed with PCA %fed through scGen~\cite{lotfollahi2019scgen} 
to obtain 50-dimensional latent codes which are clustered with $K$-Means into $8$ clusters.
%, just like in prior art~\cite{bunne2022supervised,bunne2023learning,uscidda2023monge}.
% The latent codes are clustered with $K$-Means into $8$ clusters.

\paragraph{CellOT \& CondOT.}
CellOT and CondOT are trained with the \texttt{ott-jax} package~\cite{cuturi2022optimal} for $1000$ iterations and batches of size $50$ on $\vecmu_i$, i.e., the same $8$-dimensional feature vectors (denoting a distribution of cell types over $50$ cells) used to train QontOT.
We use a cosine decay learning rate scheduler with an initial value of $0.001$ and an alpha of $0.01$, optimized with ADAM~\cite{kingma2014adam}.
CondOT uses the gaussian map initialization proposed in~\cite{bunne2022supervised}.
As CellOT and CondOT learn directly a map $f: \mathbb{R}^8 \rightarrow \mathbb{R}^8$ such that $f(\vecmu_i)=\bar{\vecnu_i}$, we use the entropically regularized sinkhorn solver~\cite{cuturi2013sinkhorn} on $(\vecmu_i, \vecnu_i)$ to obtain the transport maps and compute performance metrics on SAE and relative Frobenius norm.

\subsection{Implementation}
\label{sec:qimpl}
As mentioned in~\autoref{section-qot-formulation-i}, the fact that $U \odot \overline{U} \in \Omega_n$ is a DSM offers great flexibility in the choice of the ansatz.
In practice, we implemented two ansätze, \textit{centrosymmetric} and \textit{simple}.
Both of them have been trained in~\texttt{Qiskit} 0.43.0~\cite{Qiskit} and all experiments were performed with Qiskit's sampler class and, unless indicated otherwise (cf.~\autoref{sec:hardware}), in statevector simulation.

\paragraph{Centrosymmetric}
The centrosymmetric ansatz is our default implementation which induces a bias toward properties of centrosymmetric matrices. 
Specifically, the matrix being modelled can be divided into four quadrants such that the respective diagonals are equal and the respective off-diagonals are also equal.

\paragraph{Simple}
Instead, the simple ansatz instead is symmetric by construction and has less bias toward a specific class of unitary operators than the centrosymmetric ansatz. 
This ansatz was first formulated in~\citet{khatri2019quantum} and later refined in~\citet{madden2022best}.
Note that this ansatz implements the identity operator when all parameters are zero.\\

\FloatBarrier
\section{Neural contextual OT (NeuCOT)}
\label{section-classical-embedding}
\subsection{Methodology}
In Section \ref{subsection-tmap-embedding} we have shown that, if we produce a right stochastic matrix $\widehat{T} \in \tranpoly{\onesv{d}}{\vecnu^\prime}$,
the latter can be rescaled to a transportation map with the required initial distribution $\vecmu$.
We recall that, at inference time, the right stochastic $\widehat{T}$ is predicted by a circuit depending on the perturbation, thus the predicted transportation map results
from the rescaling of the rows of $\widehat{T}$, using the elements of the initial distribution $\vecmu$ as coefficients.
A complementary case is that of the contextual (relaxed) assignment problem, whose origin is presented in Section \ref{section:intro-OT}. The latter requires
the prediction of a DSM, and we believe that this task is hard for classical machine learning.
The former case instead, that is the one requiring a right stochastic matrix, can be shown to be practical for classical ML.
Let $M \in \mathbb{R}^{d\times d}$ and $\left[e^{M_{i, j}}\right]$ the \textit{Hadamard exponential matrix} \cite{matanalysis} of $M$,
that is the matrix resulting from the entry-wise application of the exponential mapping.
We extend the notation to a row-rescaled form, so
\begin{align}
    \label{eq:mat-map-to-rsm}
    M \mapsto& \left[\frac{e^{M_{i, j}}}{\sum_k e^{M_{i, k}}}\right],
\end{align}
then
\begin{align}
    \left[\frac{e^{M_{i, j}}}{\sum_k e^{M_{i, k}}}\right]\onesv{d}=\begin{pmatrix}
        \frac{\sum_k e^{M_{1, k}}}{\sum_k e^{M_{1, k}}}
        & \cdots &
        \frac{\sum_k e^{M_{d, k}}}{\sum_k e^{M_{d, k}}}
    \end{pmatrix}^\top = \onesv{d},
\end{align}
that is \eqref{eq:mat-map-to-rsm} maps any $M \in \mathbb{R}^{d \times d}$ to a right stochastic matrix.
Let the matrix-valued function $f_{\boldsymbol{\theta}}: \mathcal{X} \to \mathbb{R}^{d \times d}$ represent a neural network parametrised by the vector $\boldsymbol{\theta}$,
mapping the space of perturbations $\mathcal{X}$ to a $d\times d$ real matrix.
We obtain a right stochastic matrix as a function of the perturbation $\mathbf{p} \in \mathcal{X}$, so
\begin{align}
    \widehat{T} =& \left[\frac{e^{\left(f_{\boldsymbol{\theta}}(\mathbf{p})\right)_{i, j}}}
    {\sum_k e^{{\left(f_{\boldsymbol{\theta}}(\mathbf{p})\right)_{i, k}}}}\right]
    \in \tranpoly{\onesv{d}}{\vecnu^\prime}.
\end{align}
% In the language of neural networks, the latter can be interpreted as a structure with final layer a matrix of $\mathrm{SoftMax}$ activation functions.
Hence, the right stochastic matrix can be obtained through a neural network with a softmax activation in the ultimate layer. 
Finally the procedure continues as outlined in Section \ref{subsection-tmap-embedding}, that is the prediction for the transportation map
is given by $\bar{T}=D_{\vecmu} \widehat{T}$, so $\bar{T}\onesv{d}=\vecmu$ as required.
Optionally, one can make the prediction depending non-linearly on the initial distribution $\vecmu$ by re-defining $f_{\boldsymbol{\theta}}$
as a function of both the perturbation and the distribution $\vecmu$.

This is a novel, quantum-inspired approach that combines neural and contextual OT through amortized optimization.
It can be trained in a fully or weakly supervised setting, either optimizing OT plans directly (i.e., $\mathcal{L}(\bar{T},T)$) or only the push-forwarded distribution (i.e., $\mathcal{L}(\hat{T}\sharp\vecmu,\vecnu)$).
We dub this approach \textit{NeuCOT} for \underline{Neu}ral \underline{C}ontextual \underline{O}ptimal \underline{T}ransport.
Previous approaches either leverage Brenier's theorem~\yrcite{brenier1987decomposition} to recast the problem to convex regression (e.g., CellOT and CondOT~\cite{bunne2022supervised,bunne2023learning}) or use regularization~\cite{uscidda2023monge}.
In the implementation the optimization occurs over the push-forwarded measure in both cases, so unlike in our method, the OT plans can not be directly accessed.
% \TODO{Complete with argument regarding Brenier and ICNN\ldots}

\subsection{Implementation and Result}
In practice, we implement this approach with a shallow, dense neural network of two layers ($64$ and $128$ units unless indicated otherwise), a ReLU activation, a dropout ($40\%$, unless indicated otherwise) and use a MSE loss between real and predicted OT plans. 
We apply an optional residual connection of the context to the last layer of the network.
Moreover, we also apply an optional "DSM loss" to penalize deviations of the marginals from uniform ones ($\onesv{n}$).
The models have $\sim 25$k trainable parameters that are optimized with ADAM for $500$ epochs with a learning rate of $5 \mathrm{e}{-4}$.

The results in~\autoref{tab:neucot} compare NeuCOT to QontOT and the average baseline for different datasets and splitting strategies.

\begin{table}[!htb]
\centering
\begin{tabular}{ccc|cc|cc}
\textbf{Dataset} & \textbf{Data split} &  \textbf{Method} & \textbf{SAE ($\downarrow$)} & \textbf{Frob.} ($\downarrow$) & \textbf{$L_2$} ($\downarrow$) & \textbf{$R^2$} ($\uparrow$) \\ 
\hline \hline
\multirow{3}{*}{Four cell types} & \multirow{3}{*}{Random} & Average & $1.23$ & $0.83$ & $0.15$ & $0.56$ \\
 & & QontOT & $0.41$ & $0.30$ & $0.13$ & $0.54$ \\
 & & NeuCOT & $0.22$ & $0.20$ & $0.10$ & $0.73$ \\
\hline
\multirow{3}{*}{Four cell types} & \multirow{3}{*}{Extrapolation} & Average & $1.21$ & $0.84$ & $0.24$ & $0.09$ \\
 & & QontOT  & $0.53$ & $0.78$ & $0.20$ & $0.21$ \\
 & & NeuCOT & $0.32$ & $0.30$ & $0.17$ & $0.39$ \\
\hline
\multirow{3}{*}{\shortstack[c]{Known Hamiltonian \\ \\ $8 \times 8$ plans}} & \multirow{3}{*}{Random} & Average & $2.332$ & $0.883$ & $0.538$ & $0.880$ \\
 & & QontOT-$\mathcal{L}_T$ & $\textbf{0.571}$ & $\textbf{0.208}$ & $\textbf{0.085}$ & $0.892$ \\
  % & & QontOT-$\mathcal{L}_M$ & $0.630$ & $0.232$ & $0.087$ & $0.893$ \\
& & NeuCOT & $0.581$ & $0.215$ & $0.089$ & $\textbf{0.894}$ \\
 \hline
 \multirow{5}{*}{\shortstack[c]{Known Hamiltonian \\ \\ $16 \times 16$ plans}} & \multirow{5}{*}{Random} & Average & $1.317$ & $0.879$ & $0.055$ & $0.671$ \\
 & & Identity & $0.878$ & $1.042$ & $0.078$ & $0.640$ \\
 & & QontOT-$\mathcal{L}_T$ & $0.421$ & $0.833$ & $0.028$ & $\textbf{0.710}$ \\
 & & QontOT-$\mathcal{L}_M$ & $0.479$ & $0.845$ & $0.029$ & $0.699$ \\
 & & NeuCOT & $\textbf{0.400}$ & $\textbf{0.356}$ & $0.028$ & $0.708$ \\
\hline
\hline
\end{tabular}
\caption{
\textbf{Comparison of QontOT to NeuCOT.}
Means across three different runs are shown. 
All datasets use actual distributions, so $\vecmu \mathbf{1}_n = 1 = \vecnu \mathbf{1}_n$.
}
\label{tab:neucot}
\end{table}

\FloatBarrier
\section{Extended Results}

\begin{table}[!htb]
\centering
% \resizebox{1.0\columnwidth}{!}{%
\begin{tabular}{cccc|cc|cc}
\multicolumn{4}{c}{} & \multicolumn{2}{c}{\textbf{Transportation plan}} & \multicolumn{2}{c}{\textbf{Marginals}} \\
\textbf{Perturb.} & \textbf{Dist.} & \textbf{Layers} & $\mathcal{L}$ & \textbf{SAE ($\downarrow$)} & \textbf{Frob.} ($\downarrow$) & \textbf{$L_2$} ($\downarrow$) & \textbf{$R^2$} ($\uparrow$) \\ 
\hline \hline
Lin.  & $L_2$ & Ident. & -- & $1.50_{\pm 0.14}$ & $1.41_{\pm 0.12}$ & $0.69_{\pm 0.07}$ & $0.28_{\pm 0.07}$ \\
Lin.  & $L_2$ & Avg. & -- & $1.07_{\pm 0.04}$ & $0.79_{\pm 0.01}$ & $0.52_{\pm 0.04}$ & $0.27_{\pm 0.06}$ \\
% Lin. & $L_2$ & Softmax & -- & $0.21_{\pm 0.00}$ & $0.18_{\pm 0.01}$ & $0.09_{\pm 0.00}$  & $0.95_{\pm 0.03}$ \\
Lin.  & $L_2$ & QontOT  & $\mathcal{L}_T$ & $0.97_{\pm 0.08}$ & $0.70_{\pm 0.06}$ & $0.45_{\pm 0.04}$  & $0.56_{\pm 0.03}$ \\
% Lin.  & $L_2$ & QontOT-12 & $\mathcal{L}_T$ & $0.98_{\pm 0.11}$ & $0.72_{\pm 0.08}$ & $0.45_{\pm 0.05}$ & $0.53_{\pm 0.10}$ \\ 
Lin.  & $L_2$ & QontOT  & $\mathcal{L}_M$ & $0.97_{\pm 0.12}$ & $0.79_{\pm 0.15}$ & $0.41_{\pm 0.05}$  & $0.55_{\pm 0.10}$ \\
% Lin.  & $L_2$ & QontOT-12 & $\mathcal{L}_M$ & $1.02_{\pm 0.09}$ & $0.81_{\pm 0.13}$ & $0.42_{\pm 0.04}$  & $0.54_{\pm 0.08}$ \\ 
\hline
Lin. & Cos. & Ident. & -- & $1.67_{\pm 0.16}$ & $1.50_{\pm 0.13}$ & $0.69_{\pm 0.07}$ & $0.29_{\pm 0.07}$ \\
Lin. & Cos. & Avg. & -- & $1.11_{\pm 0.05}$ & $0.82_{\pm 0.02}$ & $0.52_{\pm 0.05}$ & $0.27_{\pm 0.06}$ \\
% Lin. & Cos. & Softmax & -- & $0.21_{\pm 0.01}$ & $0.17_{\pm 0.01}$ & $0.09_{\pm 0.01}$ & $0.96_{\pm 0.02}$ \\
Lin.  & Cos. & QontOT  & $\mathcal{L}_T$ & $0.97_{\pm 0.09}$ & $0.71_{\pm 0.07}$ & $0.44_{\pm 0.05}$ & $0.59_{\pm 0.02}$ \\
% Lin.  & Cos. & QontOT-12 & $\mathcal{L}_T$ &$0.99_{\pm 0.10}$ & $0.71_{\pm 0.06}$ & $0.45_{\pm 0.05}$ & $0.59_{\pm 0.17}$ \\
Lin.  & Cos. & QontOT  & $\mathcal{L}_M$ & $1.10_{\pm 0.14}$ & $0.86_{\pm 0.10}$ & $0.40_{\pm 0.06}$ & $0.59_{\pm 0.10}$ \\
% Lin.  & Cos. & QontOT-12 & $\mathcal{L}_M$ & $1.06_{\pm 0.05}$ & $0.83_{\pm 0.12}$ & $0.42_{\pm 0.05}$ & $0.60_{\pm 0.12}$ \\
\hline 
%%% WITH 100 samples
% NonLin.  & $L_2$ & Ident. & -- & $1.34_{\pm 0.11}$ & $1.30_{\pm 0.11}$ & $0.55_{\pm 0.05}$ & $0.52_{\pm 0.09}$ \\
% NonLin.  & $L_2$ & Avg. & -- & $0.95_{\pm 0.01}$ & $0.71_{\pm 0.01}$ & $0.41_{\pm 0.03}$ & $0.45_{\pm 0.18}$ \\
% NonLin. & $L_2$ & Softmax & -- & $0.24_{\pm 0.02}$ & $0.22_{\pm 0.02}$ & $0.12_{\pm 0.01}$ & $0.91_{\pm 0.00}$ \\
% % NonLin.  & $L_2$ & QontOT-6 & $\mathcal{L}_T$ & & &  & \\
% NonLin.  & $L_2$ & \TODO{QontOT-12} & $\mathcal{L}_T$  & $99.06_{\pm 0.05}$ & $0.83_{\pm 0.12}$ & $0.42_{\pm 0.05}$ & $0.60_{\pm 0.12}$ \\
% % NonLin.  & $L_2$ & QontOT-6 & $\mathcal{L}_M$ & & &  & \\
% NonLin.  & $L_2$ & QontOT-12 & $\mathcal{L}_M$  & $0.96_{\pm 0.05}$ & $0.73_{\pm 0.03}$ & $0.34_{\pm 0.02}$ & $0.56_{\pm 0.09}$ \\
% \hline
% \hline
%%% WITH 200 samples
NonLin.  & $L_2$ & Ident. & -- & $1.22_{\pm 0.05}$ & $1.19_{\pm 0.05}$ & $0.50_{\pm 0.02}$ & $0.45_{\pm 0.02}$ \\
NonLin.  & $L_2$ & Avg. & -- & $0.97_{\pm 0.02}$ & $0.72_{\pm 0.01}$ & $0.41_{\pm 0.01}$ & $0.42_{\pm 0.04}$ \\
% NonLin. & $L_2$ & Softmax & -- & $0.24_{\pm 0.02}$ & $0.22_{\pm 0.02}$ & $0.12_{\pm 0.01}$ & $0.90_{\pm 0.00}$ \\
% NonLin.  & $L_2$ & QontOT  & $\mathcal{L}_T$ & & &  & \\
NonLin.  & $L_2$ & QontOT  & $\mathcal{L}_T$  & $0.86_{\pm 0.03}$ & $0.62_{\pm 0.02}$ & $0.34_{\pm 0.01}$ & $0.47_{\pm 0.05}$ \\
% NonLin.  & $L_2$ & QontOT  & $\mathcal{L}_M$ & & &  & \\
NonLin.  & $L_2$ & QontOT  & $\mathcal{L}_M$  & $0.97_{\pm 0.06}$ & $0.77_{\pm 0.05}$ & $0.32_{\pm 0.01}$ & $0.48_{\pm 0.01}$ \\
\hline
\hline
\end{tabular}
% }
\caption{\textbf{Transportation plan prediction.}
Extended results for~\autoref{tab:syntab}, $\pm$ denotes standard deviation across three runs.
}
\label{tab:performance}
\end{table}

\begin{figure}[!htb]
    \centering
    \includegraphics[width=0.5\linewidth]{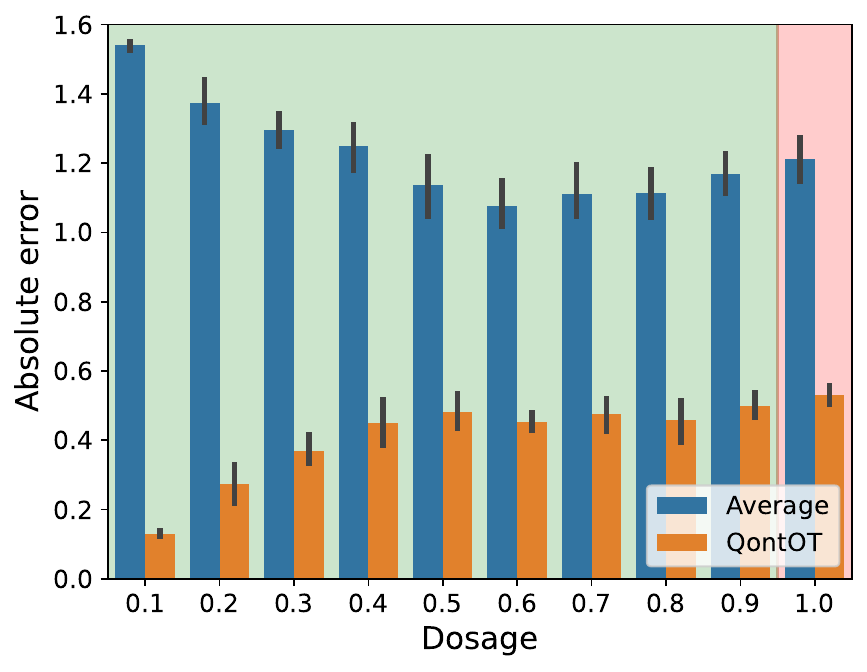}
    \caption{
    \textbf{Out-of-distribution scenario.} 
    When QontOT is evaluated on dosages outside the training data scenario (redly shaded background), the performance decreases but still remains well above the baseline.
    }
    \label{fig:extrapolation}
\end{figure}

\begin{table}[!htb]
\centering
\begin{tabular}{ccc|cc|cc}
\textbf{Dist.} & \textbf{Layers} & $\mathcal{L}$ & \textbf{SAE ($\downarrow$)} & \textbf{Frob.} ($\downarrow$) & \textbf{$L_2$} ($\downarrow$) & \textbf{$R^2$} ($\uparrow$) \\ 
\hline
  $L_2$ & 6 & $\mathcal{L}_T$ & $0.97_{\pm 0.08}$ & $0.70_{\pm 0.06}$ & $0.45_{\pm 0.04}$  & $0.56_{\pm 0.03}$ \\
  $L_2$ & 12 & $\mathcal{L}_T$ & $0.98_{\pm 0.11}$ & $0.72_{\pm 0.08}$ & $0.45_{\pm 0.05}$ & $0.53_{\pm 0.10}$ \\ 
\hline
  $L_2$ & 6 & $\mathcal{L}_M$ & $0.97_{\pm 0.12}$ & $0.79_{\pm 0.15}$ & $0.41_{\pm 0.05}$  & $0.55_{\pm 0.10}$ \\
  $L_2$ & 12 & $\mathcal{L}_M$ & $1.02_{\pm 0.09}$ & $0.81_{\pm 0.13}$ & $0.42_{\pm 0.04}$  & $0.54_{\pm 0.08}$ \\ 
\hline
  Cos. & 6 & $\mathcal{L}_T$ & $0.97_{\pm 0.09}$ & $0.71_{\pm 0.07}$ & $0.44_{\pm 0.05}$ & $0.59_{\pm 0.02}$ \\
  Cos. & 12 & $\mathcal{L}_T$ &$0.99_{\pm 0.10}$ & $0.71_{\pm 0.06}$ & $0.45_{\pm 0.05}$ & $0.59_{\pm 0.17}$ \\
\hline
  Cos. & 6 & $\mathcal{L}_M$ & $1.10_{\pm 0.14}$ & $0.86_{\pm 0.10}$ & $0.40_{\pm 0.06}$ & $0.59_{\pm 0.10}$ \\
  Cos. & 12 & $\mathcal{L}_M$ & $1.06_{\pm 0.05}$ & $0.83_{\pm 0.12}$ & $0.42_{\pm 0.05}$ & $0.60_{\pm 0.12}$ \\
\hline 
\end{tabular}
\caption{\textbf{Ablation study on number of layers in ansatz.}
Adding more layers in the ansatz and thus more parameters in the circuit does not improve performance.
Experiment performed on linear perturbation function (cf.~\autoref{tab:syntab}).
For $6$ respectively $12$ layers in the ansatz, there are $234$ respectively $456$ circuit parameters to optimize.
}
\label{tab:layerablation}
\end{table}

\begin{table}[!htb]
\centering
\begin{tabular}{c|cc|cc}
% \multicolumn{1}{c}{} & \multicolumn{2}{c}{\textbf{Transportation plan}} & \multicolumn{2}{c}{\textbf{Marginals}} \\
\textbf{Method} & \textbf{SAE ($\downarrow$)} & \textbf{Frob.} ($\downarrow$) & \textbf{$L_2$} ($\downarrow$) & \textbf{$R^2$} ($\uparrow$) \\ 
\hline \hline
Identity & $0.65$ & $0.60$ & $0.18$ & $0.46$ \\
Average & $1.23$ & $0.83$ & $0.15$ & $0.56$ \\ \hline
% Softmax & $0.22$ & $0.20$ & $0.10$ & $0.73$ \\ \hline
$\mathcal{L}_M$ & $0.59$ & $0.43$ & $0.13$ & $0.58$ \\ 
$\mathcal{L}_T$-Nevergrad & $0.41$ & $0.30$ & $0.14$ & $0.54$ \\
$\mathcal{L}_T$ & $0.41$ & $0.30$ & $0.13$ & $0.54$ \\
$\mathcal{L}_T$-AsIs & $0.45$ & $0.34$ & $0.14$ & $0.54$ \\
% $\mathcal{L}_T$-5-Anc. & $0.44$ & $0.32$ & $0.14$ & $0.54$ \\
$\mathcal{L}_T$-Simple & $0.46$ & $0.32$ & $0.13$ & $0.57$ \\
$\mathcal{L}_T$-Simple-12 & $0.55$ & $0.40$ & $0.13$ & $0.58$ \\
$\mathcal{L}_T$-Simple-12-Shared & $0.41$ & $0.30$ & $0.14$ & $0.57$ \\
$\mathcal{L}_T$-Simple-16-Shared & $0.41$ & $0.30$ & $0.14$ & $0.58$ \\
% $\mathcal{L}_T$-Trotter & $0.42$ & $0.30$ & $0.13$ & $0.59$ \\
% $\mathcal{L}_T$-Trotter-Shared & $0.45$ & $0.33$ & $0.14$ & $0.54$ \\
\hline \hline
\end{tabular}
\caption{\textbf{Ablation studies on circuit structure.}
$\mathcal{L}_T$ is the base configuration.
\textit{Simple} is an circuit constructions alternative to the base type "centrosymmetric" (cf. Appendix~\ref{sec:qimpl} for details).
"12" or "16" refers to the number of layers (base is $6$) and "Shared" defines whether the ansatz parameters are identical across layers (this is faster to optimize thus allowing deeper circuits).
"AsIs" denotes an alternative to the "atop" aggregation to produce a DSM (cf.~\autoref{subsection-tmap-embedding}).
Means across three random splits are shown.
}
\label{tab:quantablation}
\end{table}

\begin{figure*}[!htb]
    \centering
    \includegraphics[width=0.95\linewidth]{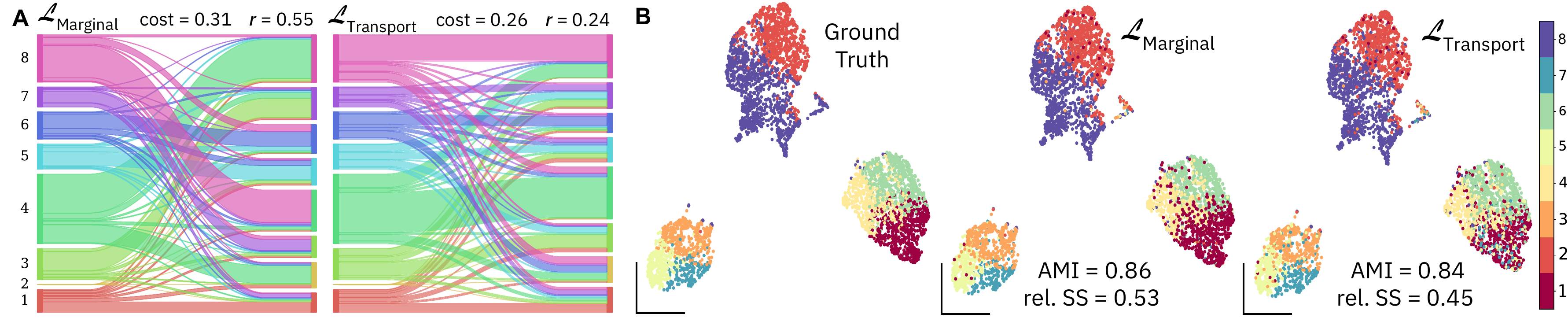}
    \caption{
    \textbf{Comparison of QontOT optimization modes.}
    \textbf{A)} Sankey plots of $\mathcal{L}_{Marginal}$ and $\mathcal{L}_{Transport}$ for a validation sample of pracinostat reveal that $\mathcal{L}_M$ learns transport maps with higher transport cost and unnecessary move of mass (see bucket $4$ and $8$).
    \textbf{B)} In contrast, the marginal performance of $\mathcal{L}_M$ is typically better, evidenced by the more discriminative cell type identification for the UMaps. 
    The higher AMI (adjusted mutual information~\cite{vinh2009information}) shows that the predicted clusters are more similar to the real ones and the relative silhouette score~\cite{rousseeuw1987silhouettes} shows that the clusters are more consistent for $\mathcal{L}_M$ than for $\mathcal{L}_T$.
    }
    \label{fig:umap}
\end{figure*}

\begin{comment}
    \subsection{Soft assignment and cell reconstruction}
\label{sec:recon}
To better capture the underlying structure of the data when generating cell labels, we employed fuzzy C-Means clustering~\cite{bezdek2013pattern} to yield a distribution of cluster assignments for each cell.
Combined with the predicted cluster label distribution $\bar{\vecnu}$ obtained from the predicted $\bar{T}_i$, this allows expression values in the original gene space to be reconstructed:
For each source cell, the mass assigned to each cluster is distributed using the corresponding row of the predicted transportation plan. These new cluster assignment vectors are summed up across all original clusters.
Last, the expression value of each gene is calculated as an average of all cluster centroids, weighted by the predicted soft cluster assignment.
%
\begin{figure}[!htb]
    \centering
    \includegraphics[width=1\linewidth]{figures/panel2.pdf}
    \caption{
    \textbf{Soft assignment.}
    \textbf{a)} Performance on different approaches of soft assignment of cells to their group.
    \textbf{b)}, \textbf{c)} Soft assignment facilitates the reconstruction of gene expression data from predicted cell label distributions. For two exemplary genes, measured and predicted gene expression across unseen dosages are shown. Expressions were averaged across all genes.
    }
    \label{fig:panel2}
\end{figure}
\end{comment}

\FloatBarrier
\section{Hardware Simulation}
\label{app:hardware-simul}

\subsection{Adhoc circuit depth optimization}
\label{sec:adhoc-circ-depth-optim}

The structure of the circuit in \autoref{fig-qc-dsm} can be optimised in terms of circuit depth when the unitary $U_p$ can be factorized as the product of two unitaries.
In our case, the latter condition is always possible since the unitary is constructed from elementary gates.
We sketch the mechanism and its proof.
Assume $U_p=U_p^{(2)} U_p^{(1)}$ (where the circuit depths of the two factors are assumed equal), and assume the state in \eqref{varphi-def} is generated by $U_p^{(1)}$, so
\begin{align}
	\ket{\varphi^{(1)}} =& \left(\idenm{2}^{\otimes m} \otimes U_p^{(1)} \otimes \idenm{2}^{\otimes n}\right)\cdot\left(\ket{b_m} \otimes \ket{b_n}\right)\notag\\
	\underset{\eqref{op-schmidt-u_p}}{=}& \sum_k \lambda_k
	\left(\idenm{2}^{\otimes m} \otimes V_k\right)\ket{b_m}
	\otimes \left(W_k \otimes \idenm{2}^{\otimes n}\right)\ket{b_n}\notag\\
	\underset{\eqref{vec-def}}{=}&
	\sum_k \lambda_k 
	\frac{\vecop\left(V_k^\top\right)}{\sqrt{2^m}} 
	\otimes \frac{\vecop\left(W_k\right)}{\sqrt{2^n}}.
\end{align}
Note that the operators $V_k$ and $W_k$ here are not the same as those in \eqref{varphi-def}, however we keep the same symbols for simplicity.
Using the identities for vectorization \eqref{vec-def} and \eqref{vec-def-ii}, we rewrite the middle expression as
\begin{align}
	\ket{\varphi^{(1)}} =& 
	\sum_k \lambda_k
	\left(V_k^{\top} \otimes \idenm{2}^{\otimes m}\right)\ket{b_m}
	\otimes \left(\idenm{2}^{\otimes n} \otimes W_k^{\top}\right)\ket{b_n},
\end{align}
Which corresponds to transposing\footnote{
	By transposition of a unitary operator we mean the complex conjugate of its adjoint, that is $\overline{U}^\dagger$.
} $U_p^{(1)}$ and changing the subsystems on which it acts. We call the new operator $K$.
Subsequently, the operator $U_p^{(2)}$, which now commutes with $K$,
is applied to the state $\ket{\varphi^{(1)}}$ with the resulting halving of the circuit depth (under the assumption stated above).
The final state is equivalent to the original one, that is
\begin{align}
    \ket{\varphi}=\left(\idenm{2}^{\otimes m} \otimes U_p^{(2)} \otimes \idenm{2}^{\otimes n}\right)\ket{\varphi^{(1)}}.
\end{align}
We illustrate the decomposition and the resulting circuit in \autoref{fig:circ-depth-halving-trick}.

\newcommand{\myugatei}{
    \ensuremath{U^{(1)}_p}
}
\newcommand{\myugateii}{
    \ensuremath{U^{(2)}_p}
}
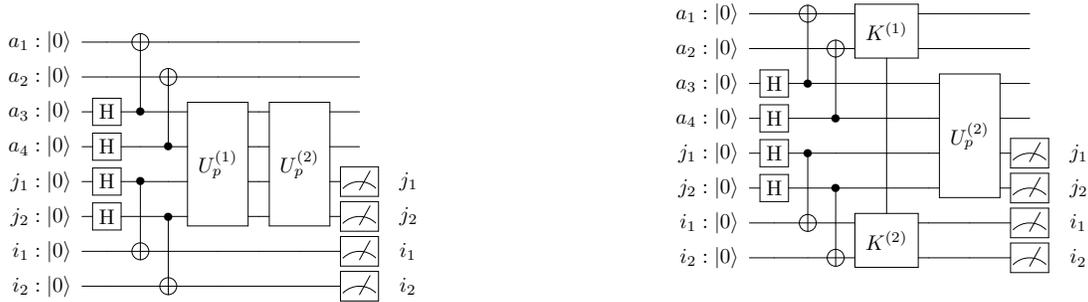
\begin{figure}[!htb]
	\centering
    \begin{subfigure}[b]{0.45\columnwidth}
    	\begin{equation*}
    		\scalebox{0.8}{
    		\Qcircuit @C=0.5em @R=0.25em @!R {
        		\nghost{ {a}_{1} : \ket{0} } & \lstick{ {a}_{1} : \ket{0} } &
    			\qw  & \targ & \qw		& \qw                                              & \qw & \qw & \qw \\
    			\nghost{ {a}_{2} : \ket{0} } & \lstick{ {a}_{2} : \ket{0} } &
    			\qw  & \qw      & \targ	& \qw                                   		   & \qw & \qw & \qw \\
                % **** start of U ****
    			\nghost{ {a}_{3} : \ket{0} } & \lstick{ {a}_{3} : \ket{0} } &
    			\gate{\mathrm{H}}  & \ctrl{-2}    & \qw 		& \multigate{3}{\myugatei} &
                \qw & \multigate{3}{\myugateii} & \qw \\
    			\nghost{ {a}_{4} : \ket{0} } & \lstick{ {a}_{4} : \ket{0} } &
    			\gate{\mathrm{H}}  & \qw      & \ctrl{-2} 	& \ghost{\myugatei}     &
                \qw & \ghost{\myugateii} & \qw \\
    			\nghost{ {j}_{1} : \ket{0} } & \lstick{ {j}_{1} : \ket{0} } &
    			\gate{\mathrm{H}} & \ctrl{2} & \qw		& \ghost{\myugatei}        &
                \qw & \ghost{\myugateii} & \meter & \rstick{j_1}\\
    			\nghost{ {j}_{2} : \ket{0} } & \lstick{ {j}_{2} : \ket{0} } &
    			\gate{\mathrm{H}} & \qw      & \ctrl{2}	& \ghost{\myugatei}		   &
                \qw & \ghost{\myugateii} & \meter & \rstick{j_2}\\
                % **** End of U ****
    			\nghost{ {i}_{1} : \ket{0} } & \lstick{ {i}_{1} : \ket{0} } &
    			\qw               & \targ    & \qw 		& \qw                                              &
                \qw & \qw & \meter & \rstick{i_1}\\
    			\nghost{ {i}_{2} : \ket{0} } & \lstick{ {i}_{2} : \ket{0} } &
    			\qw 			  & \qw      & \targ 	& \qw                               		       &
                \qw & \qw & \meter & \rstick{i_2}
    		}}
    	\end{equation*}
    	\caption{The decomposition of the circuit for $U_p(\mathbf{p};\boldsymbol{\theta})=U^{(2)}_p U^{(1)}_p$, where on the right-hand side we omit the parameters for clarity.}
    \end{subfigure}
    \hspace{1cm}
    \begin{subfigure}[b]{0.45\columnwidth}
    	\begin{equation*}
    		\scalebox{0.8}{
    		\Qcircuit @C=0.5em @R=0.25em @!R {
        		\nghost{ {a}_{1} : \ket{0} } & \lstick{ {a}_{1} : \ket{0} } &
    			\qw  & \targ & \qw		& \multigate{1}{K^{(1)}} & \qw & \qw & \qw \\
    			\nghost{ {a}_{2} : \ket{0} } & \lstick{ {a}_{2} : \ket{0} } &
    			\qw  & \qw      & \targ	& \ghost{K^{(1)}} \qwx[5] & \qw & \qw & \qw \\
                % **** start of U ****
    			\nghost{ {a}_{3} : \ket{0} } & \lstick{ {a}_{3} : \ket{0} } &
    			\gate{\mathrm{H}}  & \ctrl{-2}    & \qw 		& \qw &
                \qw & \multigate{3}{\myugateii} & \qw \\
    			\nghost{ {a}_{4} : \ket{0} } & \lstick{ {a}_{4} : \ket{0} } &
    			\gate{\mathrm{H}}  & \qw      & \ctrl{-2} 	& \qw &
                \qw & \ghost{\myugateii} & \qw \\
    			\nghost{ {j}_{1} : \ket{0} } & \lstick{ {j}_{1} : \ket{0} } &
    			\gate{\mathrm{H}} & \ctrl{2} & \qw		& \qw &
                \qw & \ghost{\myugateii} & \meter & \rstick{j_1}\\
    			\nghost{ {j}_{2} : \ket{0} } & \lstick{ {j}_{2} : \ket{0} } &
    			\gate{\mathrm{H}} & \qw      & \ctrl{2}	& \qw &
                \qw & \ghost{\myugateii} & \meter & \rstick{j_2}\\
                % **** End of U ****
    			\nghost{ {i}_{1} : \ket{0} } & \lstick{ {i}_{1} : \ket{0} } &
    			\qw               & \targ    & \qw 		& \multigate{1}{K^{(2)}} &
                \qw & \qw & \meter & \rstick{i_1}\\
    			\nghost{ {i}_{2} : \ket{0} } & \lstick{ {i}_{2} : \ket{0} } &
    			\qw 			  & \qw      & \targ 	& \ghost{K^{(2)}} &
                \qw & \qw & \meter & \rstick{i_2}
    		}}
    	\end{equation*}
    	\caption{The equivalent formulation of $U_p(\mathbf{p};\boldsymbol{\theta})$ where $U^{(1)}_p$ is substituted by the operator $K$
        which commutes with $U^{(2)}_p$. Consequently, the circuit depth of the overall gate $U_p$ is halved,
        while the total number of qubits remains the same.}
    \end{subfigure}
    \caption{
        Illustration of the adhoc compilation trick that halves the circuit depth for the unitary $U_p$.
        On the LHS we have the decomposition on the unitary $U_p$ into two factors, whereas on the RHS we represent pictorially the effect of
        the commutativity of the factors $K$ and $U_p^{(2)}$.
    }
    \label{fig:circ-depth-halving-trick}
\end{figure}

\subsection{Experimental details}

During the simulation, ansatz parameters had been optimized for $235$ iterations on a 127-qubit device (IBM Sherbrooke) available through the IBM Quantum Platform. In principle, on modern hardware every iteration can be accomplished in seconds. However, due to a very high demand on (shared) quantum computers, the actual execution time can take days. 

Result of a quantum computation is not directly observable because unlike classical bits that always take 0 or 1, qubits can exist in a superposition of states. To extract a result, one needs to measure the qubits, and this single act of measurement is called a ``shot''. Upon measurement, the superposition state collapses to one of the basis states (vectors of $0$s and $1$s) with a certain probability.

\begin{figure}[!htb]
    \centering
    \includegraphics[width=0.6\linewidth]{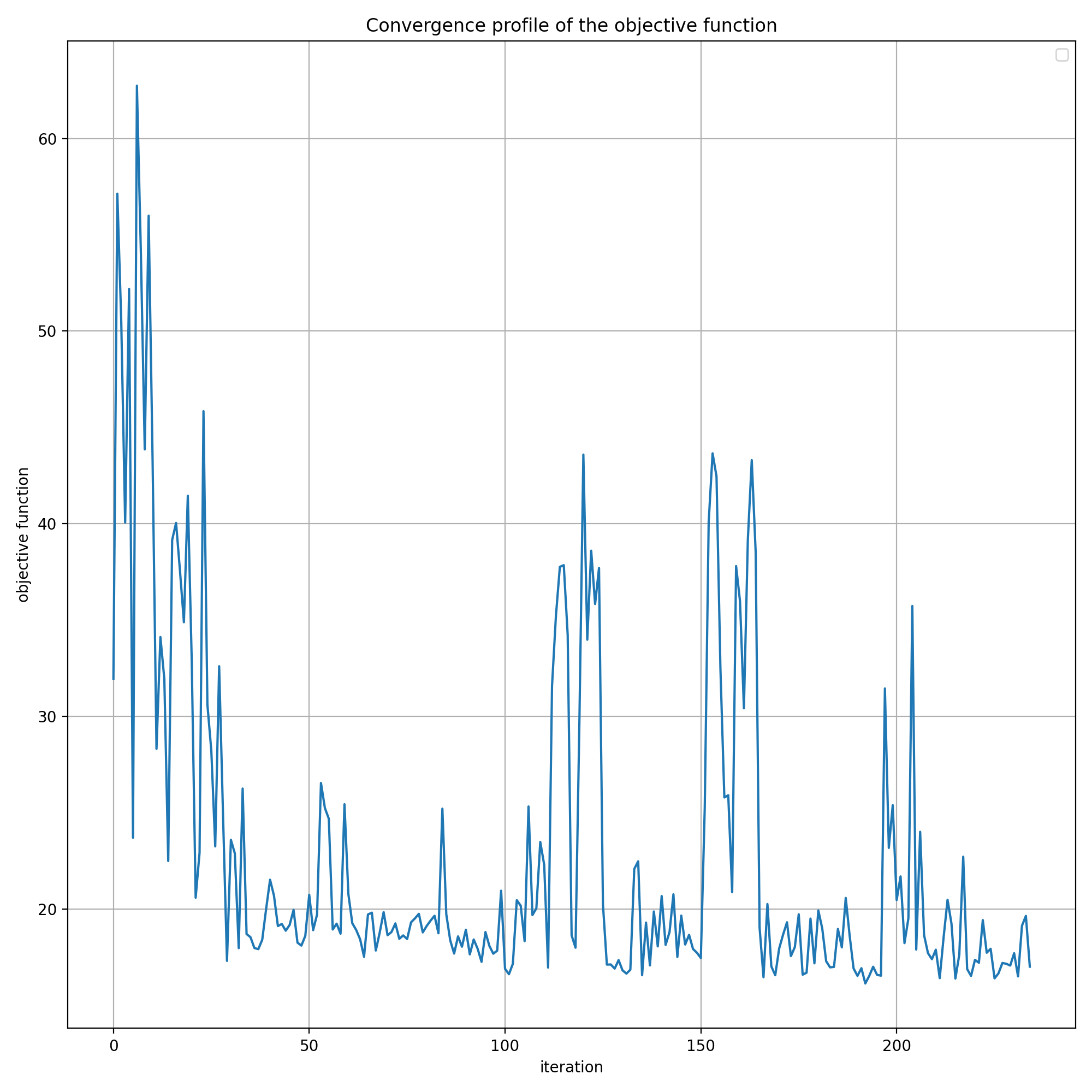}
    \vspace{-1mm}
    \caption{
    Convergence profile of the objective function while running on IBM Sherbrooke quantum device.
    }
    \label{fig:fobj-profile-hw}
\end{figure}

We did $8192$ shots per iteration and every obtained state of $0$s and $1$s was used to increment a counter of occurrences of a corresponding entry in measured DSM, just as described in the main body of this paper.
A typical picture of objective function convergence profile is shown on~\autoref{fig:fobj-profile-hw}. 

One can notice sudden spikes as the curve approaches a plateau. This happens primarily because a quantum hardware should be recalibrated from time to time. To cope with this behaviour we track the best solution (with the smallest objective function value) and report the best one rather than the last solution obtained by an optimizer. In our simulation, the best parameters had been obtained on iteration $193$.

No error mitigation was performed. It is still an open problem in quantum computing community how to suppress noise while doing state sampling. There are a number of established techniques for error mitigation when the output is a single scalar value measured on some observable. However, in this study we are interested in measuring probabilities of individual states.

\end{document}